\documentclass[conference]{IEEEtran}
\makeatletter
\def\ps@headings{%
\def\@oddhead{\mbox{}\scriptsize\rightmark \hfil \thepage}%
\def\@evenhead{\scriptsize\thepage \hfil \leftmark\mbox{}}%
\def\@oddfoot{}%
\def\@evenfoot{}}
\makeatother
\usepackage{cite}
\usepackage[final]{graphics}
\usepackage{amsmath,amssymb,amsfonts}
\usepackage{bm}
\usepackage{soul,xcolor}
\usepackage[acronym,shortcuts]{glossaries}
\usepackage{enumitem}
\usepackage{tikz}
\usepackage{stfloats}
\usepackage{amsthm}
\usepackage{siunitx}
\usepackage{comment}
\usepackage{balance}

\usepackage[caption=false]{subfig}

\newtheorem{theorem}{Theorem} 
\newtheorem{lemma}{Lemma}

\newcommand{\RM}{_\mathrm} 

\usepackage{pgfplots}
\pgfplotsset{compat=newest}
\pgfplotsset{plot coordinates/math parser=false}
\usetikzlibrary{plotmarks,arrows,decorations, patterns.meta, patterns}
\tikzdeclarepattern{
  name=hatch,
  parameters={\hatchsize,\hatchangle,\hatchlinewidth},
  bounding box={(-.1pt,-.1pt) and (\hatchsize+.1pt,\hatchsize+.1pt)},
  tile size={(\hatchsize,\hatchsize)},
  tile transformation={rotate=\hatchangle},
  defaults={
    hatch size/.store in=\hatchsize,hatch size=5pt,
    hatch angle/.store in=\hatchangle,hatch angle=0,
    hatch linewidth/.store in=\hatchlinewidth,hatch linewidth=.4pt,
  },
  code={
      \draw[line width=\hatchlinewidth] (0,0) -- (\hatchsize,\hatchsize);
  }
}

\usepackage{grffile}
\pgfplotsset{plot coordinates/math parser=false}

\definecolor{mycolor1}{rgb}{0.34667,0.53600,0.69067}%
\definecolor{mycolor2}{rgb}{0.91529,0.28157,0.28784}%
\definecolor{mycolor3}{rgb}{0.44157,0.74902,0.43216}%
\definecolor{mycolor4}{rgb}{1,0.5984,0.2}%
\definecolor{mycolor5}{rgb}{0.6769,0.4447,0.7114}%

\newlength\fwidth
\newlength\fheight
\newcommand{\pzero}[1]{p_0(#1)}
\newcommand{\puno}[1]{p_1(#1)}

\newacronym{ae}{AE}{autoencoder}
\newacronym{awgn}{AWGN}{additive white Gaussian noise}

\newacronym{cf}{CF}{channel feature}
\newacronym{cdf}{CDF}{cumulative distribution function}
\newacronym{cir}{CIR}{channel impulse response}

\newacronym{det}{DET}{detection error tradeoff}
\newacronym{dk}{DK}{delta kernel}

\newacronym{fa}{FA}{false alarm}

\newacronym{glrt}{GLRT}{generalized likelihood ratio test}
\newacronym{gmm}{GMM}{Gaussian mixture model}
\newacronym{gnss}{GNSS}{global navigation satellite system}

\newacronym{iid}{i.i.d.}{independent identically distributed}

\newacronym{kde}{KDE}{kernel density estimation}

\newacronym{llr}{LLR}{log-likelihood ratio}
\newacronym{los}{LOS}{line-of-sight}
\newacronym{lrt}{LRT}{likelihood ratio test}
\newacronym{lt}{LT}{likelihood test}
\newacronym{ls-svm}{LS-SVM}{least-square \ac{svm}}
\newacronym{ls}{LS}{least-square}

\newacronym{md}{MD}{misdetection}
\newacronym{ml}{ML}{machine learning}
\newacronym{mse}{MSE}{mean square error}
\newacronym{msgd}{MSGD}{modified SGD}

\newacronym{nn}{NN}{neural network}

\newacronym{occ}{OCC}{one-class classification}
\newacronym{osi}{OSI}{open systems interconnection}
\newacronym{ofdm}{OFDM}{Orthogonal Frequency-Division Multiplexing}
\newacronym{osvm}{OSVM}{one-class support vector machine}
\newacronym{oclssvm}{OCLSSVM}{one-class least-squares support vector machine}

\newacronym{pla}{PLA}{physical layer authentication}
\newacronym{pmd}{pmd}{probability mass distribution}
\newacronym{pdf}{pdf}{probability density function}
\newacronym{pu}{PU}{positive and unlabeled data}

\newacronym{rbf}{RBF}{radial basis function}
\newacronym{rms}{RMS}{root mean square}
\newacronym{roc}{ROC}{receiver operating characteristic}

\newacronym{sgd}{SGD}{stochastic gradient descent}
\newacronym{svm}{SVM}{support vector machine}

\newacronym{PDF}{ddadasd}{dadadadad}

\makeatletter
\newcommand{\new}[1]{{\textcolor{blue}{#1}}}

\makeatletter
\newcommand\remembertext[2]{
  \immediate\write\@auxout{\unexpanded{\global\long\@namedef{mytext@#1}{#2}}}%
  {\color{blue} #2}%
}
\newcommand\recalltext[1]{%
  \new{\ifcsname mytext@#1\endcsname
    \fontsize{10.5}{12.5}\selectfont\@nameuse{mytext@#1}%
  \else
    ``??''
  \fi
}}
\makeatother

\title{Physical Layer Authentication with Likelihood Test Using Machine Learning with Artificial Dataset}

\author{\IEEEauthorblockN{Francesco Ardizzon~{\em Member, IEEE}, and Stefano Tomasin~{\em Senior Member, IEEE}} \IEEEauthorblockA{Department of Information Engineering, University of Padova, Italy\\ \small Email: \{francesco.ardizzon, stefano.tomasin\}@unipd.it} \thanks{This work is supported by the project ISP5G+ ( CUP D33C22001300002), which is part of the SERICS program (PE00000014) under the NRRP MUR program funded by the EU-NGEU. The authors are with the Department of Information Engineering, Universit\`a degli Studi di Padova, Padua 35131, Italy. S. Tomasin is also with the National Inter-University Consortium for Telecommunications (CNIT), 43124 Parma, Italy. email:(francesco.ardizzon, stefano.tomasin)@unipd.it. }
}

\begin{document}
\maketitle

\begin{abstract}
In \ac{pla} mechanisms, a verifier decides whether a received message has been transmitted by a legitimate user or an intruder, according to some measured \acp{cf}. To design the authentication check implemented at the verifier, typically either the statistics or a dataset of features is available for the channel from the legitimate user, while no information is available when under attack.  When the statistics are known, a well-known good solution is the \ac{lt}. When a dataset is available, the decision problem is \ac{occ}, and a good understanding of the \ac{ml} techniques used for its solution is important to ensure security. Thus, in this paper, we aim at obtaining \ac{ml} \ac{pla} verifiers that operate as the \ac{lt} via \ac{nn} and least-square support vector machine (SVM) models, trained as two-class classifiers on the single-class dataset and an artificial dataset. The artificial dataset for the negative class is obtained by generating \ac{cf} vectors uniformly distributed over the domain of the legitimate class dataset. Lastly, we show that, instead, the widely used \acl{ae} classifier generally does not provide the \ac{lt}. Numerical results are provided considering \ac{pla} on both wireless and underwater acoustic channels.
\end{abstract}

\glsresetall 

\section{Introduction}\label{sec:intro}

Authentication mechanisms allow a verifier, namely Bob, to check whether a received message comes from a legitimate user, Alice, or an intruder, Trudy, impersonating Alice. In \ac{pla} schemes, the verifier extracts some \acp{cf} (i.e., the channel impulse response or the signal attenuation) from the received signal at the physical layer and checks if they are consistent with the same \acp{cf} observed in previously received authentic messages \cite{Xie2021Survey}. This solution is a valid alternative to cryptography-based authentication, which is often computationally demanding and may add overhead on the transmission, thus not suitable for energy-constrained devices or low-rate channels.

Depending on the information available to the verifier, two main authentication frameworks have been investigated. The {\em statistical framework} assumes that the \ac{pdf} (or the \ac{pmd}) of the \acp{cf} under one or both the alternative cases (authentic or not message) is available to design a {\em statistical test}; the two alternatives are called {\em hypotheses} and the \ac{pla} verification problem is called {\em hypothesis testing},~\cite[Ch. 6]{kay2009fundamentals}. More recently, a {\em \ac{ml} framework} has been explored, where the decision is made by a {\em model} that has been trained on a {\em dataset} of labeled \ac{cf} vectors obtained under one or both alternative cases; the decision alternatives are now denoted {\em classes} and the \ac{pla} verification problem {\em classification}~\cite[Ch. 5]{Goodfellow-et-al-2016}.

In typical \ac{pla} scenarios (e.g., in wireless transmissions) the verifier knows the \ac{cf} statistics or can collect \ac{cf} vectors under legitimate conditions, i.e., for the {\em null hypothesis} or the {\em positive class}, in the statistical and \ac{ml} frameworks, respectively. However, the verifier typically knows little or nothing about the intruder and thus has no \ac{pdf} (or dataset) for the observations in the {\em alternative hypothesis} (or {\em negative class}). 

In the statistical framework, without knowledge of the \ac{pdf} of the \ac{cf} under attack, the resulting \ac{pla} verification problem is called a {\em null hypothesis testing problem}. In the \ac{ml} framework without a dataset from one class, we have the {\em \ac{occ} problem}. In the statistical framework, the null hypothesis testing is based on the \ac{lt} that compares the probability of the \ac{cf} under verification with a suitably chosen threshold. 
In the \ac{ml} framework, several \ac{occ} solutions have been studied (see Section~\ref{sec:stateofart} for a detailed review), with the most relevant ones being the \ac{osvm} and the \ac{ae}.

In this paper, we aim to obtain  \ac{occ} methods that work as null-hypothesis testing. In particular, we aim at replicating the behavior of the \ac{lt} with \ac{ml} models for \ac{pla}. Indeed, knowing the behavior of a security \ac{ml} solution (thus {\em explaining} the model) helps in better understanding its potential and limits, which is crucial in security applications. We thus design \ac{ml} \ac{pla} models that, after appropriate learning, operate as the \ac{lt}, i.e., they make the same decisions. In particular, we consider multilayer perceptron \acp{nn} and \ac{ls-svm} \cite{choi2009least} models. We exploit existing results showing that both models converge to the \ac{lrt} when two labeled datasets are available. Therefore, we propose to generate an artificial dataset containing random \ac{cf} vectors uniformly distributed in the domain of \ac{cf} vectors from the dataset of legitimate \ac{cf} (positive class), to train the models. Note that the artificial dataset is used only for training, while \ac{cf} provided to the \ac{occ} at exploitation can come from any distribution of the intruder \acp{cf}. We also prove that a classifier based on the \ac{ae}  does not provide the \ac{lt}. Note that in this paper we do not present \acp{occ} with improved performance, but classifiers that converge to the \ac{lt}, which, in turn, is known to be optimal only under certain conditions. Thus, we do not compare our solution to the state-of-the-art of \ac{occ}.

In summary, the main contributions of this paper are as follows:
\begin{itemize}
    \item We train \ac{ml} models for \ac{pla} that operate as the \ac{lt}, obtaining a correspondence  between statistical and \ac{ml} frameworks. Such models are trained with an artificial dataset containing random \ac{cf} vectors, uniformly distributed in the domain of the target \ac{cf} class. 
    \item We prove that the proposed \ac{ml}-based solutions converge to \ac{lt} for sufficiently complex enough models and a sufficiently large target-class training set.
    \item We show that, in general, the \ac{ae} does not provide the \ac{lt}.
\end{itemize}

The rest of the paper is organized as follows. Section~\ref{sec:stateofart} presents the related state of the art and our contribution. Section~\ref{sec:occ_hp_testing} describes the \ac{pla} verification problem, analyzing it from both the \ac{ml} and the statistical decision theory perspectives. Section~\ref{sec:prop} introduces the \ac{lt} and the proposed learning strategy for \ac{nn}. In Section~\ref{sec:results}, we focus on two \ac{pla} contexts, i.e., radio and underwater acoustic communications, and we present the performance results also compared with the \ac{lt} and the \ac{ae}. Finally, Section~\ref{sec:conclusione} gives the conclusions.

\section{Related Literature and Main Contributions}\label{sec:stateofart}

Various tests have been proposed to assess authenticity, either based on statistics or on \ac{ml}.    
In the statistical framework, when the \acp{pdf} (or \acp{pmd}) of the \acp{cf} belonging to both hypotheses are known, the \ac{lrt} provides the minimum \ac{md} probability for a given \ac{fa} probability, as shown in the Neyman-Pearson theorem \cite{NPtest}. When the \ac{pdf} of the \acp{cf} depends on unknown parameters under one or both hypotheses, a widely used test is the \ac{glrt} \cite{kay2009fundamentals}. However, the hypothesis testing problem where the \ac{pdf} is completely unknown under one hypothesis is not much studied. A similar problem, denoted as \textit{universal outlier hypothesis testing} \cite{Li2014Universal}, aims to detect the subset of $s$ anomalies out of a set of $n$ observations.
In both \ac{glrt} and the universal outlier hypothesis testing, one or more \ac{cf} vectors from the unknown distributions are assumed to be available, and the missing \ac{pdf} (or its missing parameters) are estimated. While the \ac{lrt} has been rarely adopted for \ac{pla}, due to its strong assumption on the intruder behavior \cite{Zhang2020Physical, Xie2023Multi}, the \ac{glrt} has been considered in many papers, e.g.,  \cite{comcas,Senigagliesi2023Autoencoder,Ardizzon2024Enhancing}.
In turn, to the best of the authors' knowledge, the universal outlier detection was not considered until now for \ac{pla}.


When the statistical distribution is not known, but datasets of \acp{cf} are available, \ac{ml} solutions should be considered. For two-class classification, when labeled datasets from both classes are available during training, supervised training for classification can be applied to several models, including (deep) \ac{nn} and \ac{svm}. In \cite{brighente19} \acp{nn} and \acp{ls-svm} were shown to operate as the \ac{lrt} when the training dataset is large enough and the models are complex enough. 
In \cite{Senigagliesi2021Comparison}, the relation between statistical and \ac{ml} models in \ac{pla} has been investigated. However, the authors limited the analysis to the two-class classification case, i.e., where knowledge of either statistical descriptions or a dataset is available to Bob.

When \ac{cf} vectors are available from only one class, the \ac{occ} problem arises~\cite{Perera2021OneClass, Hoang2024Physical}. While typical solutions involve the use of \ac{ae} and the \ac{osvm} \cite{Hoang2024Physical}, several variations have been proposed in the literature. In \cite{Livi2015Entropic}, the input data is embedded in the dissimilarity space and then represented by weighted Euclidean graphs, which are used to compute the entropy of the data distribution in the dissimilarity space and obtain decision regions. In \cite{Cao2021Maximum}, it is observed that the \ac{mse} loss function for the training of \ac{osvm} is robust to the Gaussian noise but less effective against large outliers, and a robust maximum correntropy loss function is proposed. 
In particular, in~\cite{Hoang2020Detection}, an \ac{osvm} combined with K-means clustering is used for \ac{pla} to verify the authenticity of an unmanned aerial vehicle used as a relay in a wireless system. A \ac{gmm}-based predictor is used to track the channel evolution for \ac{pla} in~\cite{Qiu2019Wireless}. 
Finally, an \ac{ae}-based classifier is proposed in \cite{Hanna2021Open} to tackle the \ac{occ} problem of device fingerprinting in Wi-Fi. 




Artificial datasets for training classification models have already been considered in the literature, however, under different assumptions and with different generation techniques. In \cite{10.1007/978-3-540-87479-9_51}, a two-class classifier is used for \ac{occ}, where the dataset for the negative class is randomly generated with the {\em same} distribution as the available dataset, obtained with a \ac{pdf} estimation technique. Instead, we consider a {\em uniform} distribution to train a classifier equivalent to the \ac{lt}. In \cite{7424343}, some \ac{cf} vectors of the available dataset are considered to belong to the negative class, and the \ac{cf} vectors that have the worst fit to the one-class model are given to an expert for labeling and then used for two-class training. We do not assume any prior knowledge of the statistical distribution nor the availability of \ac{cf} vectors from the negative class. In \cite{arxiv.2206.05747}, when the negative class is described by a \ac{pdf} with unknown parameters, it is proposed to create a dataset for the two classes in binary classification (or classification with unknown parameters), instead of computing the \ac{lrt} or \ac{glrt}: in that paper, the equivalence between \ac{glrt} and the \ac{ml} techniques is supported only by a simulation campaign. On the other hand, in our paper, we assume no knowledge of the statistical distribution of the negative class (and no availability of dataset) and prove that the models converge to the \ac{lt} under certain conditions. In \cite{Oza2019OneClass}, an \ac{ae} is used to extract the features of the positive class, then a zero-mean Gaussian noise is applied in the latent space to generate \ac{cf} vectors of the negative class; datasets are then used to train an \ac{nn}. The generation of the artificial dataset is different from our approach, as we aim to obtain the \ac{lt}. Finally, generative models (see the survey \cite{Perera2021OneClass}) also include the generation of artificial datasets. In such approaches, two models are trained, the {\em discriminator} and the {\em generator}: the discriminator aims at distinguishing inputs belonging to the positive class from other inputs, while the generator aims at generating random \ac{cf} vectors that fed to the discriminator are accepted as belonging to the positive class. Even in this case, the obtained solution has not been proved to be equivalent to the \ac{lt}.

\section{System Model and PLA Verification Problem}\label{sec:occ_hp_testing}
Consider a scenario where the legitimate transmitter Alice and the malicious transmitter (intruder) Trudy are connected to receiver Bob, also the verifier, via a communication channel. Bob estimates \ac{cf} of the channel over which signals are received. In particular, Bob estimates a \ac{cf} vector $\bm{x} = [x_1, \ldots, x_M]^T$, where $^T$ denotes the transpose operator, and elements $x_j \in \mathbb R$, $j=1, \ldots, M$, are real numbers~\footnote{Here we consider real-valued vectors, but other cases can be easily accommodated in the same framework, e.g., when the vector elements are discrete or complex.}. Examples of \acp{cf} are the received signal strength indicator, the angle of arrival, and the Doppler shifts \cite{Hoang2024Physical,Senigagliesi2023Autoencoder}. 
The \ac{cf} vectors belong to a domain $\mathcal X \subseteq \mathbb R^M$, where $\mathcal X$ is also the domain of the \acp{pdf}. The \ac{cf} vector is associated to the legitimate channel, it has \ac{pdf} $\{\pzero{\bm{a}}\}$, while when the intruder is transmitting, the vector \ac{cf} has \ac{pdf} $\{\puno{\bm{a}}\}$, with $\bm{a} \in \mathcal X$. When generated according to $\{\pzero{\bm{a}}\}$, we write $\bm{x} \sim \mathcal H_0$. When $\bm{x}$ is generated from $\{\puno{\bm{a}}\}$,  we write $\bm{x} \sim \mathcal H_1$. Note that Trudy can also pre-code its signal to induce a different estimate of the \ac{cf}, and we assume here that the \ac{pdf} $\{\puno{\bm{a}}\}$ includes the effects of this precoding (see \cite{6204019} and the attack strategies described therein).

The \ac{pla} verification is the problem of deciding from which \ac{pdf}, and thus channel, the \ac{cf} vector has been generated, i.e., designing
\begin{equation}
    f(\bm{x}) \in \{{\mathcal H}_0,{\mathcal H}_1\},
\end{equation}
where $f(\cdot)$ is a deterministic transformation better detailed in the following. We assume either to either know $\pzero{\cdot}$ ({\em statistical framework}) or have a dataset of \ac{cf}s with this \ac{pdf} ({\em ML framework}); the \ac{pdf} $\puno{\cdot}$ is instead totally unknown, as it is associated to the strategy employed by Trudy. 

\subsection{Non-Separability and Decision Errors}

In typical and non-trivial \ac{pla} scenarios, while it is reasonable to consider \acp{pdf} $\{\pzero{\bm{a}}\}$ and $\{\puno{\bm{a}}\}$ to be different, there exist \ac{cf} vectors that can be observed in both classes with non-zero probability. For instance, this may be due to the fact that the measured \ac{cf} vectors are affected by (typically Gaussian) estimation errors, under both legitimate and attack conditions, which contribute to the \ac{pdf}'s spreading and overlap. 
In the \ac{ml} framework, this condition is denoted as \emph{non-separability} of the two classes.

Due to non-separability, the authentication decision is expected not to be always correct, as either \ac{fa} or \ac{md} errors may occur. In particular, an \ac{fa} occurs when a message from Alice is rejected as non-authentic, i.e.,  $f(\bm{x}) = \mathcal H_1$ while $\bm{x} \sim \mathcal H_0$. Similarly, an \ac{md} occurs when a message from Trudy is accepted as authentic, i.e., $f(\bm{x}) = \mathcal H_0$, while $\bm{x} \sim \mathcal H_1$. The corresponding \ac{fa} and \ac{md} probabilities are
\begin{equation}
\begin{split}
 P_{\rm FA}(f) & = {\mathbb P}[f(\bm{x}) ={\mathcal H}_1 | \bm{x}\sim \mathcal H_0] \\
 & =\int_{\bm{a}: f(\bm{a})={\mathcal H}_1} \pzero{\bm{a}} d\bm{a}, 
\end{split}
\end{equation} 
and
\begin{equation}
\begin{split}
 P_{\rm MD}(f) & = {\mathbb P}[f(\bm{x}) ={\mathcal H}_0 | \bm{x}\sim\mathcal H_1] \\
 & =\int_{\bm{a}: f(\bm{a})={\mathcal H}_0} \puno{\bm{a}} d\bm{a}, 
\end{split}
\end{equation}
where we have highlighted the dependency of both probabilities on the classifier or test function $f(\bm{x})$. Therefore, when designing $f(\cdot)$ both probabilities should be considered, as discussed in the following.

In the rest of this Section, we describe the design of $f(\bm{x})$ in both frameworks. In all cases, in the end, they will both compare a real value $u$, obtained from the \ac{cf} vector $\bm{x}$, with a suitable threshold $\delta$. To this end, we introduce the decision function
\begin{equation}\label{decision}
 \Delta(u, \delta) = \begin{cases} 
 \mathcal H_0 & u > \delta, \\
 \mathcal H_1 & u \leq \delta.
 \end{cases}
\end{equation}
The threshold $\delta$ is typically chosen to provide a desired \ac{fa} probability.

\subsection{LT in the Statistical Framework} 

In the statistical framework,  we say that \ac{cf} vector $\bm{x}$ belongs to one of two {\em hypotheses}: when $\bm{x} \sim \mathcal H_0$, the \ac{cf} vector belongs to the {\em null hypothesis} while when $\bm{x} \sim \mathcal H_1$, the \ac{cf} vector belongs to the {\em alternative hypothesis}. In this framework, $f(\bm{x})$ is the  {\em test function}. As we know only $\pzero{\cdot}$, we resort to the \ac{lt} to make the decision. In particular,
\begin{equation}\label{myglrt}
    f_{\rm LT}(\bm{x}) = \Delta(\log \pzero{\bm{x}}, \delta).
\end{equation}

\paragraph*{\ac{lt} as a \ac{glrt} with General Parametric \acp{pdf}} We now show that the \ac{lt} is a \ac{glrt} with specific assumptions on the \ac{pdf} of the alternative hypothesis. Suppose that the \ac{pdf} of the alternative hypothesis is parametric, i.e., $\puno{\bm{x}} = \puno{\bm{x}|\bm{\theta}}$, where $\bm{\theta}\in \Theta$ is a vector of parameters, taken from a suitable set. Considering, for example, a mixture \ac{pdf} with a large number of components, e.g., a \ac{gmm} or a \ac{kde}, such a parametric \ac{pdf} can well approximate a wide set of \acp{pdf}. The set $\Theta$ of possible parameters must be such that $0 < \puno{\bm{x},\bm{\theta}} < p_{\max}$ for any $\bm{x}$ and $\bm{\theta}$ so that we can define the \ac{glrt}
\begin{equation}\label{glrt}
     f_{\rm GLRT}(\bm{x}) = \Delta\left[\log \frac{\pzero{\bm{x}}}{\max_{\bm{\theta} \in \Theta} \puno{\bm{x},\bm{\theta}}}, \delta\right].
\end{equation}

Parametric \acp{pdf} as \ac{gmm} and \ac{kde} are typically invariant to a translation of the \ac{cf} vector $\bm{x}$, i.e., for any set of parameters $\bm{\theta}$ and for any translation vector $\bm{a}$, there exists another set of parameters $\bm{\theta}'$ such that $\puno{\bm{x},\bm{\theta}} = \puno{(\bm{x}-\bm{a}),\bm{\theta}'}$, hence we have 
\begin{equation}
    \max_{\bm{\theta} \in \Theta} \puno{\bm{x},\bm{\theta}} = p_{\max}, \quad \forall \bm{x}.
\end{equation}
In this case, the denominator in \eqref{glrt} becomes a constant, and by properly adjusting the threshold $\delta$, the \ac{glrt} is equivalent to the \ac{lt} \eqref{myglrt}.

\subsection{\ac{occ} in the \ac{ml} Framework}\label{sec:ae}

In the ML framework, the \ac{pla} verification problem is denoted as \ac{occ}. We say that \ac{cf} vector $\bm{x}$ belongs to one of two {\em classes}: when $\bm{x} \sim \mathcal H_0$, the \ac{cf} vector belongs to the {\em positive class} while when $\bm{x} \sim \mathcal H_1$, the \ac{cf} vector belongs to the {\em negative class}. In this framework, $f(\bm{x})$ is a {\em classifier}, with 
\begin{equation}\label{fML}
 f_{\rm ML}(\bm{x}) = \Delta(\mu(\bm{x},\bm{w}), \delta),
\end{equation}
where $\mu(\bm{x}, \bm{w})$ is a parametric {\em model} having as input the \ac{cf} vectors $\bm{x}$ and providing a soft real number $\mu(\bm{x}, \bm{w})$, with parameter vector $\bm{w}$. The setting of the parameters is obtained by a {\em training} using a dataset containing $N_0$ correctly labeled vector \ac{cf} vectors from the positive class, denoted as
\begin{equation}
 \mathcal D_0 = \{\bm{x}_1, \ldots, \bm{x}_{N_0}\}.
\end{equation}
What distinguishes the various \acp{occ} is the kind of used model $\mu(\cdot)$ and the way it is trained, still using the dataset $\mathcal D_0$. We describe below the considered classifiers.

\paragraph*{\Acf{ls-svm} Classifier}
The aim of \ac{ls-svm} is to find the boundary that better separates the \ac{cf} vectors of the two classes. Then during the testing phase, the user, in our case, Bob, classifies the test \ac{cf} by comparing it with the boundary in a proper domain. 

For the model of the test function \eqref{fML} we consider first an \ac{ls-svm} model $\bm{\mu}_{LS-SVM}(\bm{x}, \bm{w})$, i.e., an \ac{svm} trained using the \ac{ls} loss function, thus solving the optimization problem 
\begin{subequations}\label{eq:mse2}
\begin{align}
 \min_{\bm{w} = [\bm{w}',b]}\; & \rho_{\rm LS-SVM}({\mathcal D},\bm{w}) = \min_{\bm{w} = [\bm{w}',b]} \frac{1}{2} \bm{w}^{'\rm T}\bm{w}' + C \frac{1}{2}\sum_{n=1}^{N}e_n^2,\\
 e_n &= 1-t_n[\bm{w}^{'T} \bm{\phi}(\bm{q}_n) + b] \quad n = 1,\ldots, N\,,
\end{align}
\end{subequations}
where $\bm{w} = [\bm{w}', b]$ is a parameter vector, and $C$ is a hyper-parameter.

\paragraph*{NN Model}For the model of the test function \eqref{fML} we consider a \ac{nn} $\mu_{\rm NN}(\bm{x}, \bm{w})$, where $\bm{w}$ is the vector of parameters of the \ac{nn}. The \ac{nn} is trained to minimize the loss function $\rho_{\rm NN}({\mathcal D}, \bm{w})$, i.e.,
\begin{equation}\begin{split}
 \min_{\bm{w}}\, & \rho_{\rm NN}({\mathcal D}, \bm{w})  = \min_{\bm{w}}\, \mathbb E_{\mathcal D}[ \beta_{\rm NN}(\bm{q}, t, \bm{w}) ]\\
 & = \min_{\bm{w}}\, \left[
 \sum_{\bm{q} \in \mathcal D_0} \beta_{\rm NN}(\bm{q}, 0, \bm{w}) + \sum_{\bm{q} \in \mathcal D_1} \beta_{\rm NN}(\bm{q}, 1, \bm{w})   \right],
 \end{split}
\end{equation}
where the per-\ac{cf} vector loss function can be either the square error (for \ac{cf} vector $\bm{q}$ with label $t$)
\begin{equation}\label{eq:mse}
 \beta_{\rm NN}( \bm{q}, t, \bm{w}) =   |\mu_{\rm NN}(\bm{q}, \bm{w}) - t|^2,
\end{equation}
or the cross-entropy 
\begin{equation}\label{eq:crossentropy} \begin{split}
   \beta_{\rm NN}& ( \bm{q}, t, \bm{w}) = \\
   & t \log\mu_{\rm NN}(\bm{q}, \bm{w}) + (1-t) \log[1 - \mu_{\rm NN}(\bm{q}, \bm{w})].
\end{split}\end{equation}

\paragraph*{\Acf{ae} Classifier} 
An \ac{ae} is an unsupervised multilayer perceptron \ac{nn} trained to replicate its input to the output. The \ac{ae} can be decomposed into two sub-networks, the {\em encoder} providing an output in the {\em latent space}, and the {\em decoder}, giving as output a vector of the same size as the encoder input. The encoder \ac{nn} $f_{\rm e}(\bm{x}, \bm{w}_{\rm e})$ (with parameter vector $\bm{w}_{\rm e}$) aims at projecting the $M$-dimensional input, $\bm{x}$ into the $K$-dimensional latent space, $\bm{y}\in\mathbb{R}^K$, with $K<M$. The representation of the input in the latent space is then given as input to the decoder \ac{nn}, $f_{\rm d}(\bm{x}, \bm{w}_{\rm d})$ (with parameter vector $\bm{w}_{\rm d}$), which aims at replicating the original input, computing the reconstructed vector $\tilde{\bm{x}} = f\RM{d}(f\RM{e}(\bm{x}_n,\bm{w}_{\rm e}),\bm{w}_{\rm d})$. The \ac{ae} is trained to minimize the \ac{mse} loss function, i.e.,
\begin{equation}
\begin{split}
	 &\min_{\bm{w}}  \rho_{\rm AE}(\mathcal{D}_0, \bm{w}) =\\
	 & =\min_{\bm{w}} \frac{1}{N_0} \sum_{n=1}^{N_0}\|\bm{x}_n - f\RM{d}(f\RM{e}(\bm{x}_n,\bm{w}_{\rm e}, ),\bm{w}_{\rm d}) \|^2\,,
	\end{split}
\end{equation}
where $\bm{w} = (\bm{w}_{\rm e}, \bm{w}_{\rm d})$.
We remark that the latent space typically has a smaller dimension than the input vector, i.e., $M>K$. Thus, to replicate the input, the \ac{ae} must learn the statistical properties of the input. More details about the \ac{ae} design can be found in~\cite[Ch. 14]{Goodfellow-et-al-2016}. 

In this framework, the model used for \ac{occ} provides as output the \ac{mse} between the input \ac{cf} vector $\bm{x}$ and the \ac{ae} output $\tilde{\bm{x}}$, i.e.,
\begin{equation}
\mu_{\rm AE}(\bm{x}, \bm{w}) = \|\bm{x} - \tilde{\bm{x}} \|^2,
\end{equation}
which is then used in \eqref{fML} to obtain the \ac{ae} classifier.
The idea behind the use of an \ac{ae} for \ac{occ} is that, by training the \ac{nn} using only the $\mathcal{D}_0$ dataset, only input \ac{cf} vectors compatible with \ac{pdf} of the \ac{cf} vectors in $\mathcal{D}_0$ itself are expected to be reconstructed with low \ac{mse} during the test phase~\cite{brighente19, ribeiro2018study}.

\section{\ac{lt} with Machine Learning Models}\label{sec:prop}

We now propose models for \ac{pla} verification with suitable training that operate as the \ac{lt}. To this end, we a) show how the \ac{lt} can be described as a binary hypothesis test with a suitably defined \ac{pdf} of the alternative hypothesis; b) define training and properly selected models to be used in the \ac{ml} classifier \eqref{fML} that, using the properly selected model, operates as an \ac{lt}.

\subsection{\ac{lt} as Hypothesis Test}\label{GLRTasbinary}

In the statistical framework, when both \acp{pdf} $\pzero{\cdot}$ and $\puno{\cdot}$ are known, the uniformly most powerful test minimizing the \ac{md} probability for a given \ac{fa} probability is the \ac{lrt}, which first computes the log-likelihood ratio on the \ac{cf} vector $\bm{x}$ 
\begin{equation}\label{defgamma} 
 \Gamma(\bm{x}) = \log \frac{\pzero{\bm{x}}}{\puno{\bm{x}}}, \quad \bm{x} \in {\mathcal X},
\end{equation}
and then performs the test by comparing $\Gamma(\bm{x})$ with a threshold $\delta$, chosen to ensure the target \ac{fa} probability, i.e.,
\begin{equation}\label{lrt}
 f_{\rm LRT}(\bm{x}) = \Delta(\Gamma(\bm{x}), \delta).
\end{equation}
 
Now, we cast the \ac{lt} as a \ac{lrt}, with a properly designed alternative hypothesis \ac{pdf}, which is not the true (unknown) $\puno{\cdot}$. The following result links the \ac{lrt} of hypothesis testing with the \ac{lt} for the null hypothesis testing.
\begin{lemma}\label{lemma2}
When  the \ac{pdf} of the alternative hypothesis is constant on the domain of the null hypothesis, i.e., 
\begin{equation}\label{piC}
 \puno{\bm{a}} = u(\bm{a}) = \begin{cases}
 \frac{1}{|\mathcal X|}, & \bm{a} \in {\mathcal X},\\
 0, & {\rm otherwise},\\
\end{cases}\end{equation}
where $|\mathcal X|$ is the volume of ${\mathcal X}$, the  \ac{lt} \eqref{myglrt} is {\bf equivalent} to the \ac{lrt} \eqref{lrt}. This means that, for each threshold $\delta_1$ there exists a threshold $\delta_2$ such that
\begin{equation}\label{equivalence}
\Delta(\Gamma(\bm{x}), \delta_1) = \Delta(\pzero{\bm{x}}, \delta_2),\quad \forall \bm{x} \in \mathcal X.
\end{equation}
\end{lemma}
\begin{IEEEproof}
By inserting the definition \eqref{piC} of $u(\bm{a})$ into the log-likelihood ratio \eqref{defgamma}, we have
\begin{equation}
\Gamma(\bm{x}) = \log \pzero{\bm{x}} +\log|\mathcal X| , \quad \bm{x} \in {\mathcal X}.
\end{equation}
Considering the \ac{lrt} of \eqref{lrt}, from \eqref{decision} we have
\begin{equation}\label{derivaz}
\begin{split}
\Delta(\Gamma(\bm{x}), \delta) &= \begin{cases} 
 \mathcal H_0 & \log|\mathcal X| + \log \pzero{\bm{x}} > \delta \\
 \mathcal H_1 & \log|\mathcal X| + \log \pzero{\bm{x}} \leq \delta
 \end{cases}  \\ 
& = \begin{cases} 
 \mathcal H_0 &   \pzero{\bm{x}} > \exp[\delta - \log|\mathcal X|] \\
 \mathcal H_1 &  \pzero{\bm{x}} \leq \exp[\delta - \log|\mathcal X|]
 \end{cases} \\ 
&= \Delta(\pzero{\bm{x}}, \delta'),
\end{split}
\end{equation}
with $\delta' = \exp[\delta - \log|\mathcal X|]$. Note that the last line of \eqref{derivaz} is the \ac{lt} \eqref{myglrt}, thus the tests are equivalent in the sense of \eqref{equivalence}.
\end{IEEEproof}

Therefore, \ac{lt} can also be seen as a binary hypothesis test, where the statistic of \ac{cf} vectors under the alternative hypothesis is {\em uniform} over the null-hypothesis \ac{cf} vector domain $\mathcal X$.




\subsection{\ac{lt}-Based \ac{occ}}\label{models}

Moving now to the \ac{ml} framework, we consider here one-class classifiers implemented as follows.
\begin{enumerate}
    \item Generate an artificial dataset 
    \begin{equation}
    \mathcal{D}^\star_1 = \{\bm{v}_1, \ldots, \bm{v}_{N_1^*}\}
    \end{equation}
    of \ac{cf} vectors randomly generated according to \eqref{piC}, and for this reason, we denote it with the $^*$ mark.
    \item Train a model $\mu(\bm{x}, \bm{w})$ as a two-class classifier on the two-class labeled dataset of size $N=N_0 + N_1^*$
    \begin{equation}
    {\mathcal D} = \{{\mathcal D}_0, \mathcal{D}^\star_1\} = \{\bm{q}_1, \ldots, \bm{q}_N\},
    \end{equation}
    with labels $t_n = -1$ for \ac{cf} vectors $\bm{q}_n \in {\mathcal D}_0$ and $t_n=1$ for $\bm{q}_n \in \mathcal{D}^\star_1$.
    \item Use the trained model in the classifier \eqref{fML} to obtain the one-class classifier.
\end{enumerate}
We will show that this approach implements the \ac{lt} when using the \ac{nn} and the \ac{ls-svm}, $\mu(\bm{x},\bm{w})$, as a model.


We now show that the procedure described above allows these models to converge to the \ac{lt}.
\begin{theorem}\label{theo2}
Consider an \ac{ls-svm} $\mu_{\rm LS-SVM}(\bm{x}, \bm{w})$, trained with the \ac{ls} loss function, or a \ac{nn} $\mu_{\rm NN}(\bm{x}, \bm{w})$ trained with either the \ac{mse} or cross-entropy loss function, over a two-class labeled dataset, therefore including also the artificial dataset, i.e., $\mathcal D = \{\mathcal D_0, \mathcal{D}^\star_1\}$. When using such a model in \eqref{fML}, we obtain one-class classifiers equivalent to the \ac{lt}, when a) the training converges to the configuration minimizing the model's loss function, and b) the \ac{nn} is complex enough or the dataset $\mathcal D_0$ is large enough, the training converges to the configuration minimizing the respective loss.
\end{theorem}
\begin{IEEEproof}
First, we recall the results of \cite[Theorems 2 and 3]{brighente19}: under the hypotheses of the theorem, the multilayer perceptron \ac{nn} and the \ac{ls-svm} can converge to the global minimum. This means that, when using $\mu(\bm{x}, \bm{w})$ as a model, function \eqref{fML} implements the \ac{lrt}. 
Then, when using the artificial dataset $ \mathcal{D}^\star_1$ for the negative class, the \ac{lrt} and the \ac{lt} are equivalent, as proven in Lemma~\ref{lemma2}.

Leveraging on both the results, we can conclude that when using a \ac{ml} model, the test function \eqref{fML} converges to \ac{lt}.
\end{IEEEproof}

\subsection{On the Domain ${\mathcal X}$}

The knowledge of the domain $\mathcal X$ of the artificial dataset may not be trivial to obtain.

In typical \ac{pla} contexts, the first possibility is that we know some properties of the \ac{cf} vectors. For example, \ac{cf} vectors obtained by digital sampling an analog signal are typically clipped within an acquisition range.   

A second possibility occurs when we know that the domain of \ac{cf} vectors in the negative class is the same as those in the positive class. In this case, following the \ac{ml} approach, we can {\em learn} the domain from the dataset $\mathcal D_0$ as $\hat{\mathcal X}$. This approach works well when dataset $\mathcal D_0$ covers all points of the domain, i.e., domain $\mathcal X$ is a discrete set. When domain $\mathcal X$ is a continuous set of points, we can interpolate \ac{cf} vectors from $\mathcal D_0$ to obtain a continuous domain.
Once a $\mathcal{X}$ is identified, it is possible to introduce a normalization step, where each original \ac{cf} $x_m$ is transformed as 
\begin{equation}
    x'_m = \frac{x_m-x^{}_{m,{\rm min}}}{x_{m,{\rm max}}- x_{m,{\rm min}}}\,,
\end{equation}
where $x_{m,{\rm min}} = \min_\mathcal{X} x_m$ and $x_{m,{\rm max}} = \max_\mathcal{X} x_m$.
While this step is typically used to aid the training procedure, it also helps the artificial dataset generation, as each \ac{cf} of the artificial dataset can be generated on an $ M$-dimensional cube $[0,1] \times [0,1] \times \ldots \times [0,1] $.



A third case occurs in the absence of any knowledge of the negative class, including its domain, which is generally assumed to be different from that of the positive class. In this scenario, there are two cases to consider: a) the domain points of the positive class do not belong to the domain of the negative class, and b) the domain points of the negative class do not belong to the domain of the positive class. 

Case a) is not problematic, since if we consider $\{u(\bm{a})\}$ still uniform but over a larger domain, Lemma~\ref{lemma2} still holds, and we still get a classifier equivalent to the \ac{lt}. Instead, case b) is problematic because points of the negative class domain that do not belong to the positive class are not explored in the training phase, while they can occur in the testing phase. Since the model has not been trained on these points, its behavior is hard to predict. In this case, we can extend the domain of the artificial dataset over which the uniform \ac{cf} vectors are generated, to consider possible external points. Again, considering a larger domain for the artificial dataset even beyond the domain of the negative class does not change the resulting classifier; however, the drawbacks of the domain extensions are the need to generate a larger dataset, a slower convergence rate of the model, and a potentially more complex model (more layers and neurons) to obtain a classifier equivalent to the desired \ac{lt} also for the new input points.

We remark that, if the normalization step is adopted, normalized \ac{cf} entries having values outside the $(x_{m,{\rm min}},x_{m,{\rm max}})$ and, thus outside the $[0,1]$ interval after the normalization, can be rejected a priori and labeled as non-legitimate samples. This implicitly forces Trudy to design an attack so that the probability of falling outside $(x_{m,{\rm min}},x_{m,{\rm max}})$ is low. But this also solves the problem of case b), where Alice has to deal with values that do not belong to the domain of the positive class.

\subsection{On the \ac{ae} Classifier}
\ac{ae} classifiers have not shown good performance \cite{9087874} and several patches have been proposed. Here we confirm these deficiencies by the following result that compares the \ac{ae} classifier to the \ac{lt}.

\begin{theorem}
The \ac{ae} classifier is not equivalent to a \ac{lt}, i.e., it will make in general different classifications for the same input.
\end{theorem}

\begin{IEEEproof}
In \cite{brighente19}, it has been proven that the \ac{ae} is not equivalent to a \ac{lrt}. Thus, being the \ac{lt} a special case of \ac{lrt}, we cannot implement it with an \ac{ae}.
\end{IEEEproof}
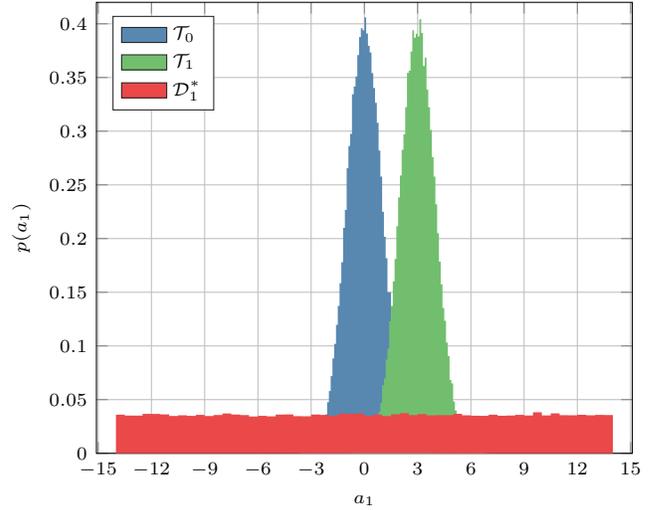
\begin{figure}
 \setlength{\fwidth}{7cm}
 \setlength{\fheight}{6cm}
 \centering
 \begin{tikzpicture}

\begin{axis}[%
width=7.125cm,
height=6cm,
at={(0,0)},
scale only axis,
xmin=-15.1,
xmax=15.1,
xtick distance=3,
ytick distance= 0.05,
xlabel={$a_1$},
ymin=0,
ymax=0.42,
yticklabel style={/pgf/number format/fixed,
                  /pgf/number format/precision=3},
ylabel={$p(a_1)$},
xmajorgrids,
ymajorgrids,
legend style={at={(0.03,0.97)}, anchor=north west, legend cell align=left, align=left, draw=white!15!black},
enlargelimits=false,title style={font=\scriptsize},xlabel style={font=\scriptsize},ylabel style={font=\scriptsize},legend style={font=\scriptsize},ticklabel style={font=\scriptsize}
]
\addplot[ybar interval, fill=mycolor1, fill opacity=0.6, draw=none, area legend] table[row sep=crcr] {%
x	y\\
-4.4	0.0001\\
-4.3	9.99999999999995e-05\\
-4.2	0\\
-4.1	9.99999999999995e-05\\
-4	0.0001\\
-3.9	0.0002\\
-3.8	0.0003\\
-3.7	0.000400000000000001\\
-3.6	0.000699999999999999\\
-3.5	0.0011\\
-3.4	0.0011\\
-3.3	0.0021\\
-3.2	0.00370000000000001\\
-3.1	0.00299999999999998\\
-3	0.00540000000000002\\
-2.9	0.00629999999999999\\
-2.8	0.00749999999999999\\
-2.7	0.012\\
-2.6	0.0163999999999999\\
-2.5	0.0223000000000001\\
-2.4	0.0246\\
-2.3	0.0335\\
-2.2	0.0362\\
-2.1	0.0476\\
-2	0.0580000000000002\\
-1.9	0.0718999999999999\\
-1.8	0.0883999999999999\\
-1.7	0.1019\\
-1.6	0.1196\\
-1.5	0.1373\\
-1.4	0.1584\\
-1.3	0.178\\
-1.2	0.21\\
-1.1	0.2267\\
-1	0.265500000000001\\
-0.9	0.2862\\
-0.8	0.2975\\
-0.7	0.3344\\
-0.6	0.3416\\
-0.5	0.351000000000001\\
-0.4	0.370899999999998\\
-0.3	0.387700000000001\\
-0.2	0.395900000000001\\
-0.100000000000001	0.393899999999998\\
0	0.406000000000001\\
0.0999999999999996	0.390999999999998\\
0.2	0.379500000000001\\
0.3	0.373299999999998\\
0.4	0.354100000000001\\
0.5	0.340200000000001\\
0.6	0.326399999999998\\
0.7	0.307800000000001\\
0.8	0.281999999999998\\
0.9	0.257700000000001\\
1	0.224900000000001\\
1.1	0.201699999999999\\
1.2	0.181700000000001\\
1.3	0.149999999999999\\
1.4	0.149800000000001\\
1.5	0.1191\\
1.6	0.104799999999999\\
1.7	0.0840000000000003\\
1.8	0.0706999999999996\\
1.9	0.0633000000000002\\
2	0.0515000000000002\\
2.1	0.0415999999999998\\
2.2	0.0322000000000001\\
2.3	0.0261999999999999\\
2.4	0.0193000000000001\\
2.5	0.0144000000000001\\
2.6	0.0111999999999999\\
2.7	0.00870000000000003\\
2.8	0.00659999999999996\\
2.9	0.00530000000000002\\
3	0.00430000000000001\\
3.1	0.00269999999999999\\
3.2	0.00320000000000001\\
3.3	0.00139999999999999\\
3.4	0.00170000000000001\\
3.5	0.000400000000000001\\
3.6	0.000400000000000001\\
3.7	0.00069999999999999\\
3.8	0.000300000000000001\\
3.9	0\\
4	0.0001\\
4.1	0.0001\\
4.2	0\\
4.3	0\\
4.4	0\\
4.5	0\\
4.6	0\\
4.7	0\\
4.8	0\\
4.9	0.0001\\
5	0.0001\\
};
\addlegendentry{$\mathcal{T}_0$}

\addplot[ybar interval, fill=mycolor3, fill opacity=0.6, draw=none, area legend] table[row sep=crcr] {%
x	y\\
-1.5	9.99999999999999e-05\\
-1.4	0\\
-1.3	0\\
-1.2	0.0002\\
-1.1	0.0002\\
-1	0.0004\\
-0.9	0.0002\\
-0.8	0.0005\\
-0.7	0.0009\\
-0.6	0.0004\\
-0.5	0.000999999999999999\\
-0.4	0.0017\\
-0.3	0.0016\\
-0.2	0.0029\\
-0.0999999999999999	0.00480000000000001\\
0	0.00529999999999999\\
0.1	0.00599999999999999\\
0.2	0.00880000000000001\\
0.3	0.0118\\
0.4	0.0135\\
0.5	0.0188\\
0.6	0.0253\\
0.7	0.0306\\
0.8	0.0384\\
0.9	0.0470000000000002\\
1	0.0633999999999999\\
1.1	0.0695999999999999\\
1.2	0.0872999999999999\\
1.3	0.0974999999999999\\
1.4	0.123\\
1.5	0.1372\\
1.6	0.1602\\
1.7	0.181\\
1.8	0.2115\\
1.9	0.238300000000001\\
2	0.2587\\
2.1	0.2827\\
2.2	0.2971\\
2.3	0.3224\\
2.4	0.354300000000001\\
2.5	0.356399999999998\\
2.6	0.374200000000001\\
2.7	0.393900000000001\\
2.8	0.387099999999998\\
2.9	0.390700000000001\\
3	0.388199999999998\\
3.1	0.404300000000001\\
3.2	0.391499999999998\\
3.3	0.360800000000001\\
3.4	0.368500000000001\\
3.5	0.338799999999998\\
3.6	0.322200000000001\\
3.7	0.301199999999998\\
3.8	0.275900000000001\\
3.9	0.257700000000001\\
4	0.231899999999999\\
4.1	0.205000000000001\\
4.2	0.178499999999999\\
4.3	0.157400000000001\\
4.4	0.1353\\
4.5	0.123399999999999\\
4.6	0.1031\\
4.7	0.0905999999999995\\
4.8	0.0683000000000002\\
4.9	0.0651000000000002\\
5	0.0481999999999997\\
5.1	0.0398000000000001\\
5.2	0.0313999999999998\\
5.3	0.0267000000000001\\
5.4	0.0193000000000001\\
5.5	0.0156999999999999\\
5.6	0.011\\
5.7	0.00899999999999995\\
5.8	0.00490000000000002\\
5.9	0.00490000000000002\\
6	0.00389999999999998\\
6.1	0.00300000000000001\\
6.2	0.00159999999999999\\
6.3	0.00160000000000001\\
6.4	0.00150000000000001\\
6.5	0.000700000000000002\\
6.6	0.000799999999999989\\
6.7	0.000400000000000001\\
6.8	0.000400000000000001\\
6.9	0.000300000000000001\\
7	0.0001\\
7.1	0\\
7.2	0.000200000000000001\\
7.3	0.000200000000000001\\
};
\addlegendentry{$\mathcal{T}_1$}

\addplot[ybar interval, fill=mycolor2, fill opacity=0.6, draw=none, area legend] table[row sep=crcr] {%
x	y\\
-14	0.03598\\
-13.5	0.03516\\
-13	0.03512\\
-12.5	0.03672\\
-12	0.03666\\
-11.5	0.03624\\
-11	0.03478\\
-10.5	0.0354\\
-10	0.03488\\
-9.5	0.0358\\
-9	0.03472\\
-8.5	0.03568\\
-8	0.0371\\
-7.5	0.0361\\
-7	0.0356\\
-6.5	0.03426\\
-6	0.0349\\
-5.5	0.03442\\
-5	0.03608\\
-4.5	0.03622\\
-4	0.03458\\
-3.5	0.03448\\
-3	0.03612\\
-2.5	0.03576\\
-2	0.03492\\
-1.5	0.03676\\
-1	0.03678\\
-0.5	0.03706\\
0	0.035\\
0.5	0.03614\\
1	0.03468\\
1.5	0.0363\\
2	0.03738\\
2.5	0.0356\\
3	0.03654\\
3.5	0.03534\\
4	0.03556\\
4.5	0.03556\\
5	0.03678\\
5.5	0.0354\\
6	0.03502\\
6.5	0.0349\\
7	0.03598\\
7.5	0.03498\\
8	0.0352\\
8.5	0.03582\\
9	0.03522\\
9.5	0.03822\\
10	0.03528\\
10.5	0.0372\\
11	0.0355\\
11.5	0.03542\\
12	0.03588\\
12.5	0.0354\\
13	0.03578\\
13.5	0.03564\\
14	0.03564\\
};
\addlegendentry{$\mathcal{D}^*_1$}

\end{axis}
\end{tikzpicture}%
 \caption{Sampling \ac{pdf} of the first entries of \ac{cf} vectors in the datasets $\{[\bm{x}]_1\}$ in the wireless AWGN Scenario: the artificially generated dataset $\mathcal{D}^\star_1$ (red) for the training phase, the $\mathcal{T}_0$ dataset of the positive-class \ac{cf} vectors (blue) for the test phase, and the  $\mathcal{T}_1$ dataset of the alternative-class \ac{cf} vectors (green)  for the test phase.}
 \label{fig:Gauss_experiments}
\end{figure}

\section{Numerical Results}\label{sec:results}

In this Section, we numerically validate the equivalence of one-class classifiers with the \ac{lt} for \ac{pla}. Then we show evidence that the \ac{ae} classifier does not provide the \ac{lt}. For the transmission, we consider both terrestrial radio communications and underwater acoustic communications.


\subsection{Channel Models}\label{sec:datasets}
We consider the two scenarios to generate datasets of \ac{cf} vectors, and each is associated with a different channel model. First, we look at the attenuations measured from terrestrial wireless channels affected by \ac{awgn}. Next, we consider a \ac{gmm} scenario, which models more general statistical distributions of \acp{cf}. For instance, we will look at features extracted from an underwater acoustic channel.
Still, note that a mixture of Gaussian variables can well fit any \ac{pdf}, thus, this scenario can be adapted to many different \ac{pla} problems.

In each scenario, datasets $\mathcal D_0$ and $\mathcal D_1^\star$ are used during the training phase. For testing, the dataset $\mathcal T = \{{\mathcal T}_0, {\mathcal T}_1\}$ is used, where ${\mathcal T}_i$, is the dataset of \ac{cf} vectors from class $\mathcal H_i$.

Sample vectors have $M$ entries and are acquired by a digital system that clips entries of the vector outside of the range $[-\zeta, \zeta]$. thus any entry $m$ of vector $\bm{x}$ such that $[\bm{x}]_m  > \zeta$ is saturated at $\zeta$, while when  $[\bm{x}]_m  < -\zeta$, the entry saturated at $-\zeta$. Let $\mathcal X_{\rm S} = [-\zeta, \zeta] \times \cdots \times [-\zeta, \zeta]$ be the domain of the clipped vectors. Therefore, for the artificial dataset $\mathcal D^\star_1$, we consider vectors with independent entries uniformly generated in the interval $[-\zeta, \zeta]$.

\begin{figure*}
    \centering
    \setlength{\fwidth}{0.45\columnwidth}
    \setlength{\fheight}{0.5\columnwidth}
    \subfloat[Number of channel taps, Alice.\label{fig:ntaps_EXP}]{\input{graphs/nTaps_estimation}}
    \subfloat[Number of channel taps, Trudy.\label{fig:ntaps_EXP_EVE}]{\definecolor{mycolor1}{RGB}{239,138,98}%
\definecolor{mycolor2}{RGB}{178,24,43}%

\begin{tikzpicture}

\begin{axis}[%
width=0.951\fwidth,
height=\fheight,
scale only axis,
xmin=4.8,
xmax=31.5,
ymin=0,
ymax=0.105,
ytick={0, 0.05, .1},
yticklabels={$0$, $0.05$, $0.1$},
xmajorgrids,
ymajorgrids,
legend style={legend cell align=left, align=left},
enlargelimits=false,title style={font=\scriptsize},xlabel style={font=\scriptsize},ylabel style={font=\scriptsize},legend style={font=\scriptsize},ticklabel style={font=\scriptsize}
]
\addplot[ybar interval, fill=mycolor1, fill opacity=0.6, draw=mycolor1, area legend] table[row sep=crcr] {%
x	y\\
6	0.0108625\\
8.4	0.0564583333333333\\
10.8	0.0966958333333333\\
13.2	0.0909166666666667\\
15.6	0.0712166666666667\\
18	0.0436875\\
20.4	0.0212666666666666\\
22.8	0.0119583333333333\\
25.2	0.0059375\\
27.6	0.00766666666666666\\
30	0.00766666666666666\\
};
\addlegendentry{Exp.}

\addplot [color=mycolor2, thick]
  table[row sep=crcr]{%
5.8774822235858	2.47402954478559e-05\\
6.0390380048736	7.77119911354873e-05\\
6.2005937861614	0.000213741045524074\\
6.2813716768053	0.000338583149279259\\
6.3621495674492	0.000520427003781521\\
6.4429274580931	0.000777353110294143\\
6.523705348737	0.00112853775183908\\
6.6044832393809	0.00159394313076788\\
6.6852611300248	0.00219134797253417\\
6.7660390206687	0.00293555565786718\\
6.8468169113126	0.00383432949976026\\
6.9275948019565	0.00488891921832035\\
7.0083726926004	0.00609128354561861\\
7.0891505832443	0.00742563028979859\\
7.1699284738882	0.00886839859889932\\
7.2507063645321	0.0103921007913499\\
7.4122621458199	0.0135639089976252\\
7.5738179271077	0.0167349457934485\\
7.65459581775159	0.0182800819032032\\
7.7353737083955	0.0197908336629737\\
7.89692948968329	0.0227251900093783\\
8.22004105225889	0.028497383558328\\
8.38159683354669	0.0314638805521881\\
8.54315261483449	0.0345191294309792\\
8.70470839612229	0.0376657551233066\\
8.86626417741009	0.0409032041609301\\
9.02781995869789	0.0442288088766425\\
9.18937573998569	0.0476315200629891\\
9.43170941191739	0.0528302048490481\\
9.91637675578079	0.0633326844334832\\
10.2394883183564	0.070301688893295\\
10.4010440996442	0.0737331816751734\\
10.562599880932	0.0771060728492117\\
10.7241556622198	0.0804003009653762\\
10.8049335528637	0.0820080634786287\\
10.8857114435076	0.0835798601153392\\
10.9664893341515	0.0851093276882189\\
11.0472672247954	0.0865857001398247\\
11.1280451154393	0.0880015973770583\\
11.2088230060832	0.0893471282732321\\
11.2896008967271	0.090619038364995\\
11.370378787371	0.0918125287729978\\
11.4511566780149	0.092930631694621\\
11.5319345686588	0.0939776317246945\\
11.6127124593027	0.0949578559027096\\
11.6934903499466	0.0958806720520222\\
11.7742682405905	0.0967491891839494\\
11.8550461312344	0.0975675517526646\\
11.9358240218783	0.0983311330434269\\
12.0166019125222	0.0990346821976367\\
12.0973798031661	0.0996642944729125\\
12.17815769381	0.100207865455889\\
12.2589355844539	0.100648777624802\\
12.3397134750978	0.10097652654807\\
12.4204913657417	0.101182691833813\\
12.5012692563856	0.101266014705036\\
12.5820471470295	0.101234935758299\\
12.6628250376734	0.101100072908721\\
12.7436029283173	0.100881379723141\\
12.8243808189612	0.100596117683413\\
12.9051587096051	0.100265494639402\\
13.0667144908929	0.0995201983112679\\
13.2282702721807	0.098705471011197\\
13.3898260534685	0.0978054260659391\\
13.4706039441124	0.097307262033155\\
13.5513818347563	0.0967684018727901\\
13.6321597254002	0.09618695607384\\
13.7129376160441	0.0955598106768214\\
13.793715506688	0.0948917636407067\\
13.9552712879758	0.0934508856241187\\
14.2783828505514	0.0904324858583117\\
14.3591607411953	0.0897140021424434\\
14.4399386318392	0.0890295724569263\\
14.5207165224831	0.0883812824650221\\
14.601494413127	0.0877687407106293\\
14.7630501944148	0.0866357320249733\\
15.1669396476343	0.0839347997399926\\
15.3284954289221	0.0827765038429575\\
15.6516069914977	0.0804301158231624\\
15.8131627727855	0.0793298807322493\\
16.1362743353611	0.0771934697790329\\
16.217052226005	0.0766228829279392\\
16.2978301166489	0.0760205619774439\\
16.3786080072928	0.0753854634736193\\
16.5401637885806	0.0740231575041577\\
17.2671648043757	0.0675953157069848\\
17.3479426950196	0.0668364681580087\\
17.4287205856635	0.0660383137408438\\
17.5094984763074	0.0651926576605142\\
17.5902763669513	0.0642980796245105\\
17.6710542575952	0.0633543206917047\\
17.7518321482391	0.0623663007321227\\
17.9133879295269	0.0602904478550172\\
18.4788331640341	0.0528108875774649\\
18.8019447266097	0.0486714133124089\\
19.1250562891853	0.0445794773893056\\
19.2866120704731	0.042585033028292\\
19.4481678517609	0.0406408969323273\\
19.6905015236926	0.0378019467887043\\
19.9328351956243	0.0350411008195408\\
20.0943909769121	0.0332694361423975\\
20.2559467581999	0.0315899304283427\\
20.3367246488438	0.0307933459304444\\
20.4175025394877	0.0300285023952291\\
20.4982804301316	0.0292951452794625\\
20.6598362114194	0.0279180134943751\\
20.8213919927072	0.0266382992680434\\
21.0637256646389	0.0248227166725989\\
21.5483930085023	0.0212685765523588\\
21.7099487897901	0.0201376897719641\\
21.8715045710779	0.0190984797670346\\
21.9522824617218	0.018626152150123\\
22.0330603523657	0.0181900436385618\\
22.1138382430096	0.0177904169081344\\
22.1946161336535	0.0174269898144779\\
22.3561719149413	0.0167936792196528\\
22.5177276962291	0.0162552042110811\\
22.6792834775169	0.0157762654096985\\
22.9216171494486	0.015123096551001\\
23.5678402745998	0.0134693925560434\\
23.7293960558876	0.0129921432952571\\
23.9717297278193	0.0122033533618833\\
24.3756191810388	0.0108379245714438\\
24.6179528529705	0.010100619512766\\
24.8602865249022	0.00943495903472069\\
25.2641759781217	0.00840124831006506\\
25.5065096500534	0.00784333902337053\\
25.7488433219851	0.00735774336363804\\
26.3142885564924	0.00629376271586324\\
26.5566222284241	0.00577151053675351\\
26.8797337909997	0.00500257829531137\\
27.2836232442192	0.00395859497148621\\
27.5259569161509	0.00334957510285605\\
27.6875126974387	0.0029833099816301\\
27.8490684787265	0.0026594617172897\\
28.0914021506582	0.00225011873936154\\
28.3337358225899	0.00191891834757385\\
28.4952916038777	0.0017413658509966\\
28.6568473851655	0.00163447631597791\\
28.7376252758094	0.00163234119646205\\
28.8184031664533	0.00168684571522704\\
28.8991810570972	0.00182126915004943\\
28.9799589477411	0.00206363444564772\\
29.060736838385	0.00244470086613191\\
29.1415147290289	0.00299436115221852\\
29.2222926196728	0.00373648502734625\\
29.3030705103167	0.00468256297560288\\
29.3838484009606	0.00582502813553631\\
29.4646262916045	0.00713153466940142\\
29.6261820728923	0.00997188937262194\\
29.7069599635362	0.0113150465139675\\
29.7877378541801	0.0124589142725213\\
29.868515744824	0.01329740627693\\
29.9492936354679	0.0137465765679963\\
30.0300715261118	0.0137578129415843\\
30.1108494167557	0.0133255682966933\\
30.1916273073996	0.0124882665324257\\
30.2724051980435	0.0113221884327608\\
30.3531830886874	0.00992932589020512\\
30.5147388699752	0.00690959036876038\\
30.5955167606191	0.00548211920115804\\
30.676294651263	0.00420642733480747\\
30.7570725419069	0.00312130279579748\\
30.8378504325508	0.00223978384599022\\
30.9186283231947	0.00155424634808554\\
30.9994062138386	0.00104296106167467\\
31.0801841044825	0.000676790692956786\\
31.1609619951264	0.000424678337591189\\
31.2417398857703	0.000257694574251843\\
31.3225177764142	0.000151203727142502\\
};
\addlegendentry{GMM fit}

\end{axis}
\end{tikzpicture}%
}
    \subfloat[Average tap power, Alice.\label{fig:avgTap_EXP}]{\input{graphs/avg_tap_pow}}
    \subfloat[Average tap power, Trudy.\label{fig:avgTap_EXP_EVE}]{\input{graphs/AvgTapPowEve}}\\
    \subfloat[\Acrshort{rms} delay, Alice.\label{fig:rms_DELAY_exp}]{\input{graphs/rms_delay_est}}
    \subfloat[\Acrshort{rms} delay, Trudy.\label{fig:rms_DELAY_exp_EVE}]{\definecolor{mycolor1}{RGB}{239,138,98}%
\definecolor{mycolor2}{RGB}{178,24,43}%

\begin{tikzpicture}

\begin{axis}[%
width=0.951\fwidth,
height=\fheight,
scale only axis,
xmin=-0.0014,
xmax=0.0167,
xtick={0.01, 0.005, 0.01, 0.015},
xticklabels={$0.01$, $0.005$, $0.01$, $0.015$},
ymin=0,
scaled ticks=false,
ymax=140,
xmajorgrids,
ymajorgrids,
legend style={at={(0.015,0.98)}, anchor=north west,legend cell align=left, align=left, draw=white!15!black, legend columns = 1},
enlargelimits=false,title style={font=\scriptsize},xlabel style={font=\scriptsize},ylabel style={font=\scriptsize},legend style={font=\scriptsize},ticklabel style={font=\scriptsize}
]
\addplot[ybar interval, fill=mycolor1, fill opacity=0.6, draw=mycolor1, area legend] table[row sep=crcr] {%
x	y\\
0	7.2375\\
0.0016	7.18125\\
0.0032	25.70625\\
0.0048	68.4875\\
0.0064	102.45\\
0.008	131.1625\\
0.0096	136.275\\
0.0112	96.1062499999999\\
0.0128	33.9625\\
0.0144	16.43125\\
0.016	16.43125\\
};
\addlegendentry{Exp.}

\addplot [color=mycolor2, thick]
  table[row sep=crcr]{%
-0.00065321259435791	0.0134447190324636\\
-0.000599073655337179	0.0239818491484698\\
-0.000544934716316448	0.0412672386268582\\
-0.000490795777324138	0.0689719322328699\\
-0.000436656838303406	0.111673793229187\\
-0.000382517899282675	0.175569837494038\\
-0.000328378960290365	0.268044666292525\\
-0.000274240021269634	0.397565175042843\\
-0.000220101082277324	0.573319739422118\\
-0.000165962143256593	0.803924600927161\\
-0.000111823204235861	1.09709563928735\\
-5.76842652435516e-05	1.45756661944446\\
-3.54532622282022e-06	1.88656191579526\\
5.05936127979112e-05	2.3807228307532\\
0.000104732551790221	2.93125108204057\\
0.000213010429831684	4.14290960160977\\
0.000321288307844725	5.37275734334537\\
0.000375427246837035	5.94357819190881\\
0.000429566185857766	6.46186450146209\\
0.000483705124878497	6.9157549629256\\
0.000537844063870807	7.29856924078516\\
0.000591983002891538	7.60913880413077\\
0.00064612194191227	7.85091552518332\\
0.00070026088090458	8.03115097710725\\
0.000754399819925311	8.15927289278807\\
0.000808538758917621	8.24546156005314\\
0.000862677697938352	8.29942046958135\\
0.000916816636959084	8.32926169976585\\
0.000970955575951393	8.34082459804122\\
0.00102509451497212	8.33738010648182\\
0.00107923345399286	8.32003173631728\\
0.00113337239298517	8.28778406964292\\
0.0011875113320059	8.23877398463239\\
0.00124165027099821	8.17075772179805\\
0.00129578921001894	8.08212614449417\\
0.00134992814903967	7.9722750902788\\
0.00140406708803198	7.84217949649207\\
0.00145820602705271	7.69455775499026\\
0.00151234496607344	7.53309736862195\\
0.00172890072209952	6.85800335786467\\
0.00178303966112026	6.70805591645862\\
0.00183717860011257	6.57414946294554\\
0.0018913175391333	6.45891617478947\\
0.00194545647815403	6.36471018591649\\
0.00199959541714634	6.29277556993429\\
0.00205373435616707	6.24483422023289\\
0.0021078732951878	6.22195559913817\\
0.00216201223418011	6.22551784667743\\
0.00221615117320084	6.25673233407079\\
0.00227029011219315	6.3169231826115\\
0.00232442905121388	6.40696607147422\\
0.00237856799023461	6.52804914716751\\
0.00243270692922692	6.68043788507345\\
0.00248684586824766	6.86417302764141\\
0.00254098480726839	7.07989428501543\\
0.0025951237462607	7.32713611368627\\
0.00264926268528143	7.60678741286799\\
0.00270340162430216	7.9189112938347\\
0.00275754056329447	8.26450064070519\\
0.0028116795023152	8.64435303081856\\
0.00286581844130751	9.05932711129282\\
0.00291995738032824	9.50955374694306\\
0.00297409631934897	9.99532810590273\\
0.00302823525834128	10.515427829741\\
0.00308237419736201	11.068414004507\\
0.00319065207537506	12.26520474228\\
0.0032989299533881	13.5685261993319\\
0.00340720783142956	14.9665762621044\\
0.0035154857094426	16.4585299202391\\
0.00362376358745564	18.0509909354662\\
0.00373204146546868	19.7522752263296\\
0.00384031934351015	21.5698171429601\\
0.00394859722152319	23.5142959223153\\
0.00405687509953623	25.6105872308633\\
0.00416515297757769	27.8992392344711\\
0.00421929191657	29.1299998638656\\
0.00427343085559073	30.4236933496932\\
0.00438170873360377	33.2057689864452\\
0.00448998661161681	36.2274953854864\\
0.00459826448965828	39.4289965536895\\
0.00508551494073117	54.1519904005669\\
0.00530207069678568	60.3940622606272\\
0.00573518220886626	72.7113664756504\\
0.00589759902590004	77.3995070595166\\
0.00600587690391308	80.3845078115216\\
0.00606001584293381	81.7892995422916\\
0.00611415478192612	83.1211914038066\\
0.00616829372094685	84.3718392992076\\
0.00622243265993916	85.5400948581387\\
0.00627657159895989	86.6279127701052\\
0.00633071053798062	87.6426963223588\\
0.00638484947697293	88.5928212169113\\
0.00649312735501439	90.3439888235811\\
0.00681796098905352	95.334166400102\\
0.00692623886709498	97.2145813016479\\
0.00703451674510802	99.258666835964\\
0.00730521144015484	104.560452690747\\
0.00741348931816788	106.519377341518\\
0.00773832295223542	112.173317257921\\
0.00784660083024846	114.280048627565\\
0.00795487870828993	116.519447001628\\
0.0081172955253237	119.916704861288\\
0.00822557340333674	122.031743474731\\
0.00827971234232905	123.00800616908\\
0.00833385128134978	123.919247477546\\
0.00838799022037051	124.768262892106\\
0.00844212915936282	125.558243757296\\
0.00855040703740428	127.010557757508\\
0.00871282385440963	129.117831690299\\
0.0088211017324511	130.652075296035\\
0.00892937961046414	132.327233234065\\
0.00909179642749791	134.910383688233\\
0.00914593536649022	135.721678316506\\
0.00920007430551095	136.478287834836\\
0.00925421324453168	137.170508970984\\
0.00930835218352399	137.786061544166\\
0.00936249112254472	138.322726201489\\
0.00941663006156546	138.77462427165\\
0.00947076900055777	139.144247274806\\
0.0095249079395785	139.433563960434\\
0.00957904687859923	139.647463296885\\
0.00963318581759154	139.791961939625\\
0.00968732475661227	139.874331224674\\
0.00974146369560458	139.903282861839\\
0.00979560263462531	139.884634314897\\
0.00984974157364604	139.830417481604\\
0.00990388051263835	139.74342972037\\
0.00995801945165908	139.631985661379\\
0.0100121583906798	139.494615090057\\
0.0100662973296721	139.333621826776\\
0.0101204362686929	139.145305564861\\
0.0101745752076852	138.923875303841\\
0.0102287141467059	138.66381344266\\
0.0102828530857266	138.357625177674\\
0.0103369920247189	138.000098890052\\
0.0103911309637397	137.583002577913\\
0.0104452699027604	137.10596059459\\
0.0104994088417527	136.563534223673\\
0.0105535477807734	135.957963480831\\
0.0106076867197942	135.286037808976\\
0.0106618256587865	134.550752320696\\
0.0107159645978072	133.752988525606\\
0.0107701035367995	132.892780042784\\
0.0108242424758203	131.973702894932\\
0.010878381414841	130.99523258171\\
0.0109325203538333	129.962002467934\\
0.0110407982318748	127.735988025962\\
0.0111490761098878	125.305797849943\\
0.0112573539879008	122.659792281022\\
0.0113114929269216	121.242960326514\\
0.0113656318659139	119.753635268653\\
0.0114197708049346	118.185405270602\\
0.0114739097439553	116.529902829305\\
0.0115280486829477	114.786439057899\\
0.0116363265609607	111.031011232375\\
0.0117446044390022	106.955270725279\\
0.0118528823170152	102.638038397767\\
0.012015299134049	95.8963314999228\\
0.0123942717071088	79.8056363278075\\
0.0127191053411764	66.0673700928553\\
0.0129356610972309	57.1305893586561\\
0.013152216853257	48.3961882568256\\
0.0133146336702907	42.0804526239451\\
0.0134229115483038	38.0992249533998\\
0.0135311894263452	34.3846599604522\\
0.0136394673043583	30.9865363493278\\
0.0136936062433506	29.4145145540716\\
0.0137477451823713	27.9277185169423\\
0.013801884121392	26.525977292673\\
0.0138560230603844	25.2055183149023\\
0.0139101619994051	23.9629209988935\\
0.0139643009384258	22.7916956449347\\
0.0140184398774181	21.6853296911234\\
0.0141267177554312	19.6344170018517\\
0.0142349956334726	17.7406198833511\\
0.0143432735114857	15.9575473190578\\
0.0144515513894987	14.2587257938604\\
0.0145056903285194	13.4565335414481\\
0.0145598292675402	12.7059496689861\\
0.0146139682065325	12.0305415923295\\
0.0146681071455532	11.4629827251602\\
0.0147222460845455	11.0431204341002\\
0.0147763850235663	10.8180731312229\\
0.014830523962587	10.8382049544785\\
0.0148846629015793	11.1538579505371\\
0.0149388018406	11.8064928232966\\
0.0149929407796208	12.8215987453036\\
0.0150470797186131	14.1966751309215\\
0.0151012186576338	15.8937063179416\\
0.0151553575966261	17.8325508971566\\
0.0152636354746676	21.9071095927018\\
0.0153177744136599	23.7032073368969\\
0.0153719133526806	25.0969068728469\\
0.0154260522917014	25.9292643946617\\
0.0154801912306937	26.0877192991704\\
0.0155343301697144	25.5221152112008\\
0.0155884691087067	24.2536282662617\\
0.0156426080477274	22.3706779396437\\
0.0156967469867482	20.0163298731405\\
0.0158050248647612	14.6063214169696\\
0.0158591638037819	11.9058257640267\\
0.0159133027427742	9.40364181661232\\
0.015967441681795	7.19605117219106\\
0.0160215806208157	5.33464553782986\\
0.016075719559808	3.83093623311777\\
0.0161298584988288	2.66469030574856\\
0.0161839974378211	1.79524809310158\\
0.0162381363768418	1.17140314790939\\
0.0162922753158625	0.740242458658486\\
0.0163464142548548	0.453014199036971\\
0.0164005531938756	0.268481222891637\\
};
\addlegendentry{GMM fit}
\end{axis}
\end{tikzpicture}%
}
    \subfloat[Smoothed received power, Alice.\label{fig:sm_power_exp}]{\input{graphs/sm_power_est}}
    \subfloat[Smoothed received power, Trudy.\label{fig:sm_power_exp_EVE}]{\definecolor{mycolor1}{RGB}{239,138,98}%
\definecolor{mycolor2}{RGB}{178,24,43}%

\begin{tikzpicture}

\begin{axis}[%
width=0.951\fwidth,
height=\fheight,
scale only axis,
xmin=2.15,
xmax=7.87,
ymin=0,
ymax=0.47,
xmajorgrids,
ymajorgrids,
legend style={legend cell align=left, align=left, draw=white!15!black, legend columns = 1},
enlargelimits=false,title style={font=\scriptsize},xlabel style={font=\scriptsize},ylabel style={font=\scriptsize},legend style={font=\scriptsize},ticklabel style={font=\scriptsize}
]
\addplot[ybar interval, fill=mycolor1, fill opacity=0.6, draw=mycolor1, area legend] table[row sep=crcr] {%
x	y\\
2.4	0.0904807692307692\\
2.92	0.298730769230769\\
3.44	0.429711538461538\\
3.96	0.458326923076923\\
4.48	0.371846153846154\\
5	0.159384615384615\\
5.52	0.0442884615384615\\
6.04	0.0307692307692308\\
6.56	0.0135384615384615\\
7.08	0.026\\
7.6	0.026\\
};
\addlegendentry{Exp.}

\addplot [color=mycolor2, thick]
  table[row sep=crcr]{%
2.27191209428405	0.000110889552192184\\
2.30693522959473	0.000380702561278312\\
2.32444679725007	0.000666280453133439\\
2.34195836490541	0.00112246599820587\\
2.35946993256075	0.00182667825409588\\
2.37698150021609	0.00287205594358841\\
2.39449306787143	0.00436733696723035\\
2.41200463552677	0.00642664946981153\\
2.42951620318211	0.00916162984714131\\
2.44702777083744	0.0126610758798824\\
2.46453933849278	0.0169849659492236\\
2.48205090614812	0.0221390438495757\\
2.49956247380346	0.028079515995616\\
2.5170740414588	0.0347041651037907\\
2.53458560911414	0.0418669745746438\\
2.55209717676948	0.0493893355639905\\
2.6046318797355	0.072365481464681\\
2.62214344739084	0.0797138572629326\\
2.63965501504618	0.0867995937317092\\
2.65716658270152	0.0936253459072889\\
2.67467815035686	0.10023701176934\\
2.70970128566753	0.113082434226184\\
2.74472442097821	0.12589801871283\\
2.77974755628889	0.139078890097172\\
2.81477069159957	0.152748519715229\\
2.84979382691025	0.166819358442672\\
2.90232852987627	0.188367792922102\\
2.95486323284229	0.210238793680727\\
2.98988636815297	0.225029402706082\\
3.04242107111898	0.247559303096121\\
3.11246734174034	0.277758027713076\\
3.14749047705102	0.292439074185917\\
3.16500204470636	0.299531616124179\\
3.1825136123617	0.306404351495506\\
3.20002518001704	0.313019140027852\\
3.21753674767238	0.319354819605839\\
3.23504831532772	0.325401495713971\\
3.25255988298306	0.331168224625936\\
3.27007145063839	0.336672089376583\\
3.28758301829373	0.341947051922626\\
3.30509458594907	0.347026079977715\\
3.34011772125975	0.356744070291489\\
3.37514085657043	0.366040202752459\\
3.41016399188111	0.374971512786797\\
3.42767555953645	0.379272858159524\\
3.44518712719179	0.38343618216699\\
3.46269869484713	0.3874486484672\\
3.48021026250247	0.391296730395028\\
3.49772183015781	0.394986002168368\\
3.53274496546848	0.401945007827137\\
3.56776810077916	0.408479801872067\\
3.60279123608984	0.414673124085186\\
3.62030280374518	0.417611892906554\\
3.63781437140052	0.420420099423724\\
3.65532593905586	0.423055327037103\\
3.6728375067112	0.425504643432032\\
3.69034907436654	0.427745158599271\\
3.70786064202188	0.429776541567414\\
3.72537220967722	0.431615350391543\\
3.74288377733256	0.433285251793559\\
3.77790691264324	0.43627248129929\\
3.84795318326459	0.441929502278009\\
3.88297631857527	0.444985141063443\\
3.97053415685197	0.452850881032062\\
3.98804572450731	0.454213090997372\\
4.00555729216265	0.455411548502192\\
4.02306885981799	0.456412149186575\\
4.04058042747333	0.457192835864048\\
4.05809199512866	0.457749327435473\\
4.07560356278401	0.458094949661987\\
4.11062669809468	0.458306860809804\\
4.1631614010607	0.458335010621656\\
4.18067296871604	0.458513379951712\\
4.19818453637138	0.458839343771558\\
4.23320767168206	0.459900597272942\\
4.26823080699274	0.461157991954197\\
4.28574237464808	0.461650469516407\\
4.30325394230342	0.461925176261761\\
4.32076550995876	0.461899210912255\\
4.33827707761409	0.461507084194738\\
4.35578864526943	0.460707555370018\\
4.37330021292477	0.459488907052405\\
4.39081178058011	0.457860998581499\\
4.40832334823545	0.455859975477562\\
4.42583491589079	0.453532991258193\\
4.44334648354613	0.450935527769179\\
4.46085805120147	0.448125293563432\\
4.47836961885681	0.445150466395921\\
4.51339275416749	0.438821030274187\\
4.53090432182283	0.435482374202097\\
4.54841588947817	0.432009405615019\\
4.56592745713351	0.428387957295936\\
4.58343902478885	0.424582387230251\\
4.60095059244419	0.420584194777439\\
4.61846216009952	0.416366884030619\\
4.63597372775486	0.411928457900438\\
4.6534852954102	0.407255738514981\\
4.67099686306554	0.402341608989189\\
4.68850843072088	0.397166103645533\\
4.70601999837622	0.391712556886648\\
4.72353156603156	0.385953398065545\\
4.7410431336869	0.379874514464122\\
4.75855470134224	0.373459505028886\\
4.77606626899758	0.366715373595245\\
4.79357783665292	0.35965245829884\\
4.81108940430826	0.352303433173404\\
4.8286009719636	0.344700237787627\\
4.84611253961894	0.336888113933879\\
4.88113567492961	0.320792213660838\\
4.93367037789563	0.296039783701795\\
4.96869351320631	0.279570413681414\\
5.00371664851699	0.263465947418722\\
5.02122821617233	0.255614299051276\\
5.05625135148301	0.240323257541354\\
5.09127448679369	0.225448543511543\\
5.16132075741504	0.19613693492086\\
5.19634389272572	0.181654654722523\\
5.2313670280364	0.167537021618308\\
5.24887859569174	0.160686037523462\\
5.26639016334708	0.15400370516442\\
5.28390173100242	0.147498144575484\\
5.30141329865776	0.141177182572538\\
5.3189248663131	0.135039389374897\\
5.33643643396844	0.129084251530035\\
5.35394800162378	0.123303483804934\\
5.37145956927912	0.117694902107876\\
5.38897113693446	0.11225187690526\\
5.4064827045898	0.10697491325984\\
5.42399427224514	0.101865740338821\\
5.44150583990047	0.0969333821889959\\
5.45901740755581	0.0921861463730158\\
5.47652897521115	0.0876395297129022\\
5.49404054286649	0.0833079011971511\\
5.51155211052183	0.0791990528679287\\
5.52906367817717	0.0753218555575135\\
5.54657524583251	0.0716726383518749\\
5.56408681348785	0.068249387914709\\
5.58159838114319	0.0650406982463965\\
5.59910994879853	0.0620380658880597\\
5.61662151645387	0.0592293897265748\\
5.63413308410921	0.0566095152137613\\
5.65164465176455	0.0541702604720165\\
5.66915621941989	0.0519108161180082\\
5.68666778707523	0.0498275395550651\\
5.70417935473056	0.0479189947976497\\
5.7216909223859	0.0461791983883728\\
5.73920249004124	0.0446029884795429\\
5.75671405769658	0.0431783748893739\\
5.79173719300726	0.0407365938628761\\
5.82676032831794	0.0387364848775995\\
5.86178346362862	0.0370602306766159\\
5.8968065989393	0.0356086842586709\\
5.94934130190532	0.0337476543515782\\
5.98436443721599	0.0327424896145949\\
6.01938757252667	0.0320004471545667\\
6.05441070783735	0.0315684070921991\\
6.08943384314803	0.0314247870189925\\
6.14196854611405	0.0315348646387417\\
6.21201481673541	0.0316840969263428\\
6.26454951970142	0.0314645043958981\\
6.31708422266744	0.0308639962760457\\
6.40464206094414	0.0293933046626131\\
6.4746883315655	0.028068391407845\\
6.50971146687618	0.0271755393299307\\
6.54473460218685	0.0259962621347745\\
6.57975773749753	0.0245127707876591\\
6.63229244046355	0.0218915910598021\\
6.71985027874025	0.0173905327580277\\
6.77238498170627	0.0149172671922537\\
6.82491968467228	0.012680410270014\\
6.8774543876383	0.010756892292064\\
6.92998909060432	0.0091584157609752\\
7.01754692888102	0.00686684185274711\\
7.12261633481305	0.00431665609729137\\
7.15763947012373	0.00363631178524493\\
7.19266260543441	0.00315263129188192\\
7.22768574074509	0.00291579142450082\\
7.24519730840043	0.00295350481104961\\
7.26270887605577	0.00315882795846623\\
7.28022044371111	0.00360978671063794\\
7.29773201136645	0.00441399325252512\\
7.31524357902179	0.00570700966506443\\
7.33275514667713	0.00764382210194015\\
7.35026671433246	0.0103787009851057\\
7.3677782819878	0.0140346785850403\\
7.38528984964314	0.0186650808989963\\
7.40280141729848	0.0242118186726952\\
7.42031298495382	0.030474015441655\\
7.4553361202645	0.0435853938482671\\
7.47284768791984	0.0493656060900651\\
7.49035925557518	0.0538551344891154\\
7.50787082323052	0.0565619351702598\\
7.52538239088586	0.0571701554314075\\
7.5428939585412	0.0555987092707211\\
7.56040552619654	0.0520168146956603\\
7.57791709385188	0.0468123826274223\\
7.59542866150722	0.0405213878135502\\
7.63045179681789	0.0270125253270157\\
7.64796336447323	0.0208014679205935\\
7.66547493212857	0.0154052883845557\\
7.68298649978391	0.0109719416515777\\
7.70049806743925	0.00751507981563115\\
7.71800963509459	0.0049500367004125\\
7.73552120274993	0.00313556316093511\\
7.75303277040527	0.00191006132417559\\
7.77054433806061	0.00111891991394764\\
7.78805590571595	0.000630327314790513\\
};
\addlegendentry{GMM fit}
\end{axis}
\end{tikzpicture}%
}
    \caption{Experimental (light) versus fitted \ac{gmm} model (dark) for Alice (blue) and Trudy (red).}
    \label{fig:experimental_vs_fitted}
\end{figure*}

\paragraph*{\bf Wireless AWGN Channel}
In the AWGN scenario, the transmitter (i.e., Alice or Trudy) sends a signal, while Bob, upon reception, measures the raw channel frequency response. Next, after proper thresholding, Bob extracts the amplitude of the first $M$ channel taps, which are considered to be the \acp{cf} and collected in vector $\bm{x}$. 
More in detail, 
\begin{itemize}
	\item we consider the transmission of a signal, from which $M=4$ channel tap amplitudes are measured, 
	\item the $M$ measurements are affected by independent \ac{awgn},
	\item we assume that Alice's position is public while Bob's and Trudy's positions are secret,
	\item before performing the test, Bob normalizes each measurement by subtracting the mean (which is a function of the distance between Alice and Bob) and dividing by the standard deviation.  
\end{itemize}
We remark that the mean attenuation can be easily computed by Bob since it is a function of the \ac{los} path loss, which, in turn, is a function of the distance Alice-Bob. 
A similar model has been adopted in the context of drone authentication in \cite{droniGLOBECOM}.

After normalization, $\bm{x}$ has a clipped multivariate Gaussian \ac{pdf} with unitary variance per entry and independent entries, i.e., for $i=0$ (positive class) and $1$ (negative class), entry $j$ of \ac{cf} vector $\bm{x}$ has \ac{pdf} 
\begin{equation}
p_i(a_j) = \begin{cases}
f_{\rm G}(a_j), & a_j \in (-\zeta, \zeta),\\
\delta_{\rm D}(a_j-\zeta) \int_{a \geq \zeta} f_{\rm G}(a_j) da
 & a_j= \zeta, \\
\delta_{\rm D}(a_j+\zeta)\int_{a \leq-\zeta} f_{\rm G}(a_j) da
 & a_j = - \zeta, \\
0, & {\rm otherwise},
\end{cases} 
\end{equation}
where  $\delta_{\rm D}(\cdot)$ is the Dirac delta function and
\begin{equation}\label{eq:distGauss}
    f_{\rm G}(a) = \frac{1}{\sqrt{2\pi}} \exp \left( {-\frac{|a - \bm{\gamma}_i|^2}{2}}\right)\;.
\end{equation}
Thus due to the normalization for the positive class (thus for \ac{cf} vectors of datasets $\mathcal D_0$ and $\mathcal T_0$) the mean is $\bm{\gamma}_0 = 0 \cdot\bm{1}_4$, (here $\bm{1}_4 = [1, 1, 1, 1]^T$). On the other hand, we assume Trudy to be in a position such that the \ac{cf} vector has again unitary variance, but mean in the test phase $\gamma_1 = 3\cdot\bm{1}_4$.

Fig.~\ref{fig:Gauss_experiments} shows the sampling \ac{pdf} of the first element of \ac{cf} vectors, from the testing datasets $\mathcal T_0$ and  $\mathcal T_1$, and from the artificial dataset $\mathcal D^\star_1$.  
\begin{figure}
    \setlength{\fwidth}{9cm}
    \setlength{\fheight}{6cm}
    \centering
    \input{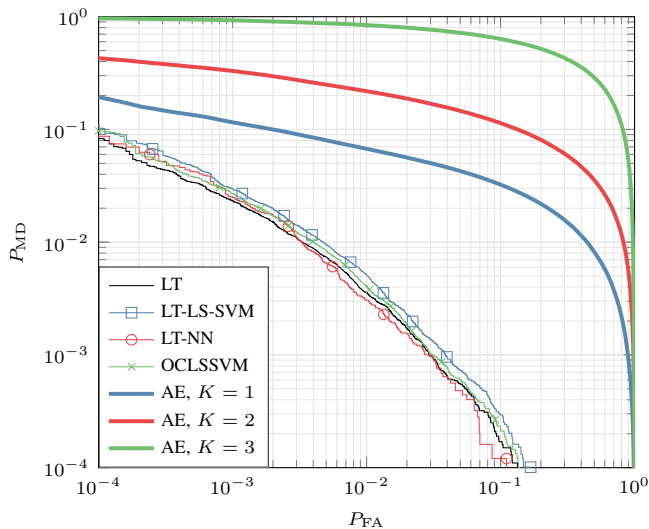}
    \caption{\ac{det} curves for the \ac{awgn} Scenario for various classifiers and the \ac{lt}.}
    \label{fig:gauss_vs_AE}
\end{figure}

\paragraph*{\bf Underwater Acoustic Channel} 
We now consider a scenario where the distribution of the \acp{cf} can be fit using a \ac{gmm}. This is appropriate for the features of an underwater acoustic communication channel, as discussed in \cite{paperUCOMMS}, where a \ac{gmm} model obtained via \ac{kde} is used to model the underwater acoustic \acp{cf} extracted from the power-delay profile. Fig.~\ref{fig:experimental_vs_fitted} shows an example of an experimental versus estimated \ac{gmm} model considering the number of channel taps, the \ac{rms} delay, the average tap power, and the smoothed power. Dataset and experimental setting are detailed in \cite{comcas}. Another use of the \ac{gmm} to fit the \acp{cf} extracted from an underwater acoustic channel can be found in \cite{Khalid2020Physical} to model the signals' angle of arrival. 

Thus $\bm{x}$ is a vector of $M=4$ independent entries. Under hypothesis $\mathcal H_i$, $i=0,1$, entry $m$ is a Gaussian mixture of $\nu_{m,i}$ components with means $\{\gamma_{m,i,j}, j=1, \ldots, \nu_{m,i}\}$, and mixing probabilities $\{q_{m,i,j}, j=1, \ldots, \nu_{m,i}\}$. Each entry $m$ of \ac{cf} vector $\bm{x}$ has \ac{pdf} for class $i$   
\begin{equation}
p_{m,i}(a) = \begin{cases} g_{m,i}(a)  & a \in (-\zeta,\zeta), \\
\delta_{\rm D}(a-\zeta)\int_{b\geq\zeta} g_{m,i}(b) db & a = \zeta\\
\delta_{\rm D}(a+\zeta)\int_{b\leq\zeta} g_{m,i}(b) db  & a = -\zeta \\
0 & {\rm otherwise},
\end{cases}
\end{equation}
where 
\begin{equation}
    g_{m,i}(a) = \frac{1}{\sqrt{2\pi}}\sum_{j=1}^{\nu_{m,i}} q_{m,i,j} \exp \left( {-\frac{|a - \gamma_{m,i,j}|^2}{2}} \right) \,.
\end{equation}

\subsection{Classifier's Architecture and Training}
For all approaches, the test phase of the \ac{pla} protocol (from which their performance is assessed) operates on a test dataset of $25000$ \ac{cf} vectors coming from both the positive and the negative classes. We now detail the parameters used for each classifier.


\paragraph*{\ac{lt}-Based \ac{nn} (LT-NN) Classifier} We design the \ac{nn} with $7$ layers with $40$, $32$, $24$, $16$, $8$, $4$, and $1$ neurons, respectively; all the neurons have sigmoid activation functions. The training lasted for 5 epochs; the one-class training and validation datasets have $60000$ and $15000$ \ac{cf} vectors, respectively. The artificial dataset used for training has $60000$ \ac{cf} vectors.

\paragraph*{\ac{lt}-Based \ac{ls-svm} (LT-LS-SVM) Classifier} As kernel function we use the \ac{rbf}. Due to the computational cost of the SVM approach, we used a training dataset containing 5000 \ac{cf} vectors, with $\alpha = 2.3$. 


\paragraph*{Autoencoder (\ac{ae}) Classifier} We consider a linear \acp{ae}, with $4$ neurons in both input and output layers, and linear activation functions. In the hidden layer, we have instead either $K=1$, $2$, or $3$ neurons, still with linear activation functions. Weights are initialized randomly. The model has been trained with 5 epochs and the datasets of the LT-NN classifier.

We remark that, in both cases, the parameter $\alpha$ was tuned by exhaustive search.

\begin{figure}
 \setlength{\fwidth}{9cm}
 \setlength{\fheight}{6cm}
 \centering
 \input{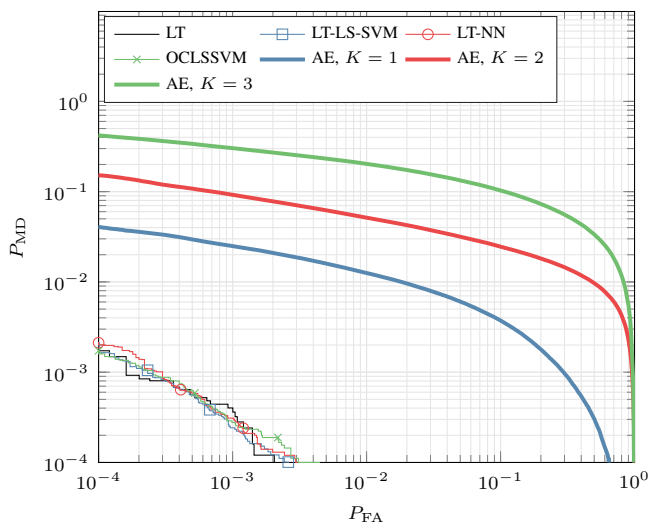}
 \caption{\ac{det} curves for the Mixture Scenario for various classifiers and the \ac{lt}.}
 \label{fig:GMM_GLRT_AE}
\end{figure}

\subsection{Performance Evaluation}
To evaluate the performance of the \ac{pla} classifiers on the AWGN and Gaussian mixture scenarios, mapping the terrestrial radio and the underwater communication context, respectively, we consider the \ac{det} curves, showing the \ac{md} probability as a function of the \ac{fa} probability achieved during the test phase.

\paragraph*{AWGN Scenario} Fig.~\ref{fig:gauss_vs_AE} shows the \ac{det} for the various considered solutions in the Gaussian Scenario. We note that the \ac{lt}-based classifiers perform as the \ac{lt}, as expected. 
We note that the \ac{ae} classifier does not perform as well as the \ac{lt} and, for the considered dataset, shows a worse performance. Indeed, the \ac{ae} classifier performance improves as $K$ decreases, i.e., with a more compact latent space.



%

\paragraph*{Underwater Communication Scenario} Fig.~\ref{fig:GMM_GLRT_AE} shows the \ac{det} for the \ac{pla} classifiers and the \ac{lt} in the underwater acoustic communication scenario, which may also translate to more complex scenarios. In this case, all classifiers and the \ac{lt} are better performing than in the AWGN Scenario, due to the more marked differences between the \acp{pdf} of the \ac{cf} vectors of the two classes. Also in this case, we observe that all \ac{lt}-based classifiers have a similar performance and show a \ac{det} very close to that of the \ac{lt}. 



\section{Conclusions}\label{sec:conclusione}

We considered the \ac{occ} problem in \ac{pla} aimed at identifying classifiers that learn the \ac{lt}, based on the availability of only the legitimate dataset. We have solutions where either a \ac{nn} or a \ac{ls-svm} model is trained as a two-class classifier using an artificially generated dataset. We have investigated the conditions under which these models converge to the \ac{lt}, then confirmed by numerical results on a Gaussian and a \ac{gmm}, modeling respectively a wireless \ac{awgn} channel, an underwater acoustic channel. 
Additionally, we have shown that the \ac{ae} one-class classifier does not converge in general to the \ac{lt}. 
\balance

\bibliographystyle{IEEEtran}
\bibliography{IEEEabrv,biblio}

\begin{thebibliography}{10}
\providecommand{\url}[1]{#1}
\csname url@samestyle\endcsname
\providecommand{\newblock}{\relax}
\providecommand{\bibinfo}[2]{#2}
\providecommand{\BIBentrySTDinterwordspacing}{\spaceskip=0pt\relax}
\providecommand{\BIBentryALTinterwordstretchfactor}{4}
\providecommand{\BIBentryALTinterwordspacing}{\spaceskip=\fontdimen2\font plus
\BIBentryALTinterwordstretchfactor\fontdimen3\font minus \fontdimen4\font\relax}
\providecommand{\BIBforeignlanguage}[2]{{%
\expandafter\ifx\csname l@#1\endcsname\relax
\typeout{** WARNING: IEEEtran.bst: No hyphenation pattern has been}%
\typeout{** loaded for the language `#1'. Using the pattern for}%
\typeout{** the default language instead.}%
\else
\language=\csname l@#1\endcsname
\fi
#2}}
\providecommand{\BIBdecl}{\relax}
\BIBdecl

\bibitem{Xie2021Survey}
N.~Xie, Z.~Li, and H.~Tan, ``A survey of physical-layer authentication in wireless communications,'' \emph{IEEE Commun. Surv. Tutor.}, vol.~23, no.~1, pp. 282--310, Dec. 2020.

\bibitem{kay2009fundamentals}
S.~Kay, \emph{Fundamentals Of Statistical Processing, Volume 2: Detection Theory}.\hskip 1em plus 0.5em minus 0.4em\relax Pearson Education, 2009.

\bibitem{Goodfellow-et-al-2016}
I.~Goodfellow, Y.~Bengio, and A.~Courville, \emph{Deep Learning}.\hskip 1em plus 0.5em minus 0.4em\relax MIT Press, 2016, \url{http://www.deeplearningbook.org}.

\bibitem{choi2009least}
Y.-S. Choi, ``Least squares one-class support vector machine,'' \emph{Pattern Recognit. Lett.}, vol.~30, no.~13, pp. 1236--1240, Oct. 2009.

\bibitem{NPtest}
J.~{Neyman} and E.~S. {Pearson}, ``On the problem of the most efficient tests of statistical hypotheses,'' \emph{Philosophical Trans. of the Royal Society of London Series A}, vol. 231, pp. 289--337, Jan. 1933.

\bibitem{Li2014Universal}
Y.~Li, S.~Nitinawarat, and V.~V. Veeravalli, ``Universal outlier hypothesis testing,'' \emph{IEEE Trans. on Info. Theory}, vol.~60, no.~7, pp. 4066--4082, Apr. 2014.

\bibitem{Zhang2020Physical}
P.~Zhang, T.~Taleb, X.~Jiang, and B.~Wu, ``Physical layer authentication for massive {MIMO} systems with hardware impairments,'' \emph{IEEE Trans. Wirel. Commun.}, vol.~19, no.~3, pp. 1563--1576, Mar. 2020.

\bibitem{Xie2023Multi}
N.~Xie, M.~Sha, T.~Hu, and H.~Tan, ``Multi-user physical-layer authentication and classification,'' \emph{IEEE Trans. Wirel. Commun.}, vol.~22, no.~9, pp. 6171--6184, Sept. 2023.

\bibitem{comcas}
L.~Bragagnolo, F.~Ardizzon, N.~Laurenti, P.~Casari, R.~Diamant, and S.~Tomasin, ``Authentication of underwater acoustic transmissions via machine learning techniques,'' in \emph{Proc. of COMCAS}, 2021, pp. 255--260.

\bibitem{Senigagliesi2023Autoencoder}
L.~Senigagliesi, G.~Ciattaglia, and E.~Gambi, ``Autoencoder based physical layer authentication for {UAV} communications,'' in \emph{Proc. of Vehicular Technology Conference (VTC2023-Spring)}, 2023, pp. 1--6.

\bibitem{Ardizzon2024Enhancing}
F.~Ardizzon, L.~Crosara, S.~Tomasin, and N.~Laurenti, ``Enhancing spreading code signal authentication in {GNSS}: a {GLRT}-based approach,'' in \emph{Proc. of ICL-GNSS}, 2024, pp. 1--6.

\bibitem{brighente19}
A.~Brighente, F.~Formaggio, G.~M. Di~Nunzio, and S.~Tomasin, ``Machine learning for in-region location verification in wireless networks,'' \emph{IEEE J. Sel. Areas Commun.}, vol.~37, no.~11, pp. 2490--2502, Nov. 2019.

\bibitem{Senigagliesi2021Comparison}
L.~Senigagliesi, M.~Baldi, and E.~Gambi, ``Comparison of statistical and machine learning techniques for physical layer authentication,'' \emph{IEEE Trans. Inf. Forensics Secur.}, vol.~16, pp. 1506--1521, Oct. 2021.

\bibitem{Perera2021OneClass}
\BIBentryALTinterwordspacing
P.~Perera, P.~Oza, and V.~M. Patel, ``One-class classification: {A} survey,'' 2021. [Online]. Available: \url{https://arxiv.org/abs/2101.03064}
\BIBentrySTDinterwordspacing

\bibitem{Hoang2024Physical}
T.~M. Hoang, A.~Vahid, H.~D. Tuan, and L.~Hanzo, ``Physical layer authentication and security design in the machine learning era,'' \emph{IEEE Commun. Surv. Tutor.}, pp. 1--1, Feb. 2024.

\bibitem{Livi2015Entropic}
L.~Livi, A.~Sadeghian, and W.~Pedrycz, ``Entropic one-class classifiers,'' \emph{IEEE Trans. Neural Netw. Learn. Syst.}, vol.~26, no.~12, pp. 3187--3200, Dec. 2015.

\bibitem{Cao2021Maximum}
J.~Cao, H.~Dai, B.~Lei, C.~Yin, H.~Zeng, and A.~Kummert, ``Maximum correntropy criterion-based hierarchical one-class classification,'' \emph{IEEE Trans. Neural Netw. Learn. Syst.}, vol.~32, no.~8, pp. 3748--3754, Aug. 2021.

\bibitem{Hoang2020Detection}
T.~M. Hoang, N.~M. Nguyen, and T.~Q. Duong, ``Detection of eavesdropping attack in {UAV}-aided wireless systems: {U}nsupervised learning with one-class {SVM} and {K}-means clustering,'' \emph{IEEE Wirel. Commun. Lett.}, vol.~9, no.~2, pp. 139--142, Feb. 2020.

\bibitem{Qiu2019Wireless}
X.~Qiu, T.~Jiang, S.~Wu, C.~Jiang, H.~Yao, M.~H. Hayes, and A.~Benslimane, ``Wireless user authentication based on {KLT} and {G}aussian mixture model,'' in \emph{Proc. of WCNC}, 2019, pp. 1--5.

\bibitem{Hanna2021Open}
S.~Hanna, S.~Karunaratne, and D.~Cabric, ``Open set wireless transmitter authorization: Deep learning approaches and dataset considerations,'' \emph{IEEE Trans. Cogn. Commun. Netw.}, vol.~7, no.~1, pp. 59--72, Mar. 2021.

\bibitem{10.1007/978-3-540-87479-9_51}
K.~Hempstalk, E.~Frank, and I.~Witten, ``One-class classification by combining density and class probability estimation,'' in \emph{Proc. of ECML PKDD}, 2008.

\bibitem{7424343}
V.~Barnabé-Lortie, C.~Bellinger, and N.~Japkowicz, ``Active learning for one-class classification,'' in \emph{Proc. of ICMLA}, 2015, pp. 390--395.

\bibitem{arxiv.2206.05747}
T.~Diskin, U.~Okun, and A.~Wiesel, ``Learning to detect with constant false alarm rate,'' in \emph{Proc. of SPAWC}, 2022, pp. 1--5.

\bibitem{Oza2019OneClass}
P.~Oza and V.~M. Patel, ``One-class convolutional neural network,'' \emph{IEEE Signal Process. Lett.}, vol.~26, no.~2, pp. 277--281, Feb. 2019.

\bibitem{6204019}
P.~Baracca, N.~Laurenti, and S.~Tomasin, ``Physical layer authentication over {MIMO} fading wiretap channels,'' \emph{IEEE Trans. Wirel. Commun.}, vol.~11, no.~7, pp. 2564--2573, 2012.

\bibitem{ribeiro2018study}
M.~Ribeiro, A.~E. Lazzaretti, and H.~S. Lopes, ``A study of deep convolutional auto-encoders for anomaly detection in videos,'' \emph{Pattern Recognit. Lett.}, vol. 105, no.~C, pp. 13--22, Apr. 2018.

\bibitem{9087874}
M.~Ribeiro, M.~Gutoski, A.~E. Lazzaretti, and H.~S. Lopes, ``One-class classification in images and videos using a convolutional autoencoder with compact embedding,'' \emph{IEEE Access}, vol.~8, pp. 86\,520--86\,535, May 2020.

\bibitem{droniGLOBECOM}
\BIBentryALTinterwordspacing
F.~Ardizzon, D.~Salvaterra, M.~Piana, and S.~Tomasin, ``Energy-based optimization of physical-layer challenge-response authentication with drones,'' in \emph{arXiv}, 2024. [Online]. Available: \url{https://arxiv.org/abs/2405.03608}
\BIBentrySTDinterwordspacing

\bibitem{paperUCOMMS}
F.~Ardizzon, R.~Diamant, P.~Casari, and S.~Tomasin, ``Machine learning-based distributed authentication of {UWAN} nodes with limited shared information,'' in \emph{Proc. UComms}, 2022, pp. 1--5.

\bibitem{Khalid2020Physical}
M.~Khalid, R.~Zhao, and N.~Ahmed, ``Physical layer authentication in line-of-sight underwater acoustic sensor networks,'' in \emph{Proc. of OCEANS}, 2020, pp. 1--5.

\end{thebibliography}

\end{document}